%% file: main.tex
\documentclass[runningheads]{llncs}

\usepackage[mobile]{eccv}

\usepackage{eccvabbrv}

\usepackage{graphicx}
\usepackage{booktabs}

\usepackage[accsupp]{axessibility}  

\usepackage[pagebackref,breaklinks,colorlinks,citecolor=eccvblue]{hyperref}

\usepackage{orcidlink}

\usepackage{kotex}
\usepackage{xcolor}
\usepackage{ulem}
\usepackage{comment}
\usepackage{hhline}
\usepackage{booktabs}
\usepackage{multirow}
\usepackage{colortbl}
\usepackage{boldline}
\usepackage{adjustbox}

\newcommand{\myparagraph}[1]{\vspace{2pt}\noindent{\bf #1}}
\begin{document}

\title{Your Vision-Language-Action Model Already Has Attention Heads For Path Deviation Detection} 

\titlerunning{VLA Models Already Detect Path Deviation}

\author{
Jaehwan Jeong \inst{1,2}\orcidlink{0009-0006-1894-5266} 
\and
Evelyn Zhu \inst{2}
\and 
Jinying Lin \inst{2}
\and
Emmanuel Jaimes \inst{2}
\and \\
Tuan-Anh Vu \inst{2}\orcidlink{0000-0002-8872-0875}
\and 
Jungseock Joo \inst{2,3}
\and 
Sangpil Kim \inst{1,\dagger}\orcidlink{0000-0002-7349-0018}
\and
M. Khalid Jawed \inst{2,\dagger}\orcidlink{0000-0003-4661-1408}
}

{\renewcommand{\thefootnote}{}\footnotetext{$^\dagger$ Co-corresponding authors.}}

\authorrunning{J. Jeong et al.}

\institute{
$^1$Korea University\quad
$^2$University of California, Los Angeles\quad
$^3$NVIDIA
}

\maketitle

\vspace{-3pt}
\begin{abstract}
    Vision-Language-Action (VLA) models have demonstrated strong potential for predicting semantic actions in navigation tasks, demonstrating the ability to reason over complex linguistic instructions and visual contexts. However, they are fundamentally hindered by visual-reasoning hallucinations that lead to trajectory deviations. Addressing this issue has conventionally required training external critic modules or relying on complex uncertainty heuristics.
    In this work, we discover that monitoring a few attention heads within a frozen VLA model can accurately detect path deviations without incurring additional computational overhead. We refer to these heads, which inherently capture the spatiotemporal causality between historical visual sequences and linguistic instructions, as Navigation Heads. Using these heads, we propose an intuitive, training-free anomaly-detection framework that monitors their signals to detect hallucinations in real time. Surprisingly, among over a thousand attention heads, a combination of just three is sufficient to achieve a 44.6\,\% deviation detection rate with a low false-positive rate of 11.7\,\%.
    Furthermore, upon detecting a deviation, we bypass the heavy VLA model and trigger a lightweight Reinforcement Learning (RL) policy to safely execute a shortest-path rollback. By integrating this entire detection-to-recovery pipeline onto a physical robot, we demonstrate its practical robustness. All source code will be publicly available.

\end{abstract}

\input{tex/1_introduction}
\input{tex/2_related}

\input{tex/3_method}
\input{tex/4_experiment}
\input{tex/5_conclusion}

\clearpage
\section*{Acknowledgements}
We acknowledge financial support from the National Institute of Food \& Agriculture of the U.S. Department of Agriculture (Grant No. 2024-67021-42528) and the NVIDIA Academic Grants Program. 
This work was also supported by the Culture, Sports, and Tourism R\&D Program through a Korea Creative Content Agency grant funded by the Ministry of Culture, Sports and Tourism in 2024 (International Collaborative Research and Global Talent Development for the Development of Copyright Management and Protection Technologies for Generative AI, No. RS-2024-00345025), as well as an Institute of Information \& Communications Technology Planning \& Evaluation (IITP) grant funded by the Korean government (MSIT) (No. RS-2019-II190079, Artificial Intelligence Graduate School Program, Korea University).

\bibliographystyle{splncs04}
\bibliography{main}

\clearpage
\input{tex/supplementary}


\end{document}

%% file: tex/1_introduction.tex
\section{Introduction}
\label{sec:intro} 

Driven by recent advances in Large Vision-Language Models (LVLMs), vision-language navigation has progressed from discrete graph-based agents to Transformer based policies that can reason over long instructions and rich visual observations~\cite{anderson2018vision, fried2018speaker, tan2019learning, hong2021vln, chen2021history, chen2022think}. More recently, Vision-Language-Action (VLA) models have extended this paradigm to continuous environments by directly mapping sequential visual observations and language instructions to executable navigation actions~\cite{krantz2020beyond}. This formulation is attractive because it removes the need for predefined topological constraints and enables agents to operate in open-world settings using unified multimodal reasoning~\cite{zhang2024navid, cheng2025navila, wang2024lookahead, marafioti2025smolvlm}.

Although VLA models have demonstrated significant potential for predicting semantic actions, they are fundamentally limited by inheriting the chronic hallucination issues of their Large Language Model (LLM) backbones. As the agent accumulates longer visual histories while following multi-step natural-language instructions, failures in visual reasoning and language grounding can lead to trajectory drift, incorrect sub-goal selection, and ultimately catastrophic path deviations~\cite{tang2025seeing, chen2024halc}. Prior work has highlighted the importance of detecting uncertainty and initiating backtracking when an agent loses its path~\cite{an2024etpnav, chen2021topological, ma2019regretful, wang2023dreamwalker}.

\begin{figure*}[!t]
\centering
\includegraphics[width=0.9\textwidth]{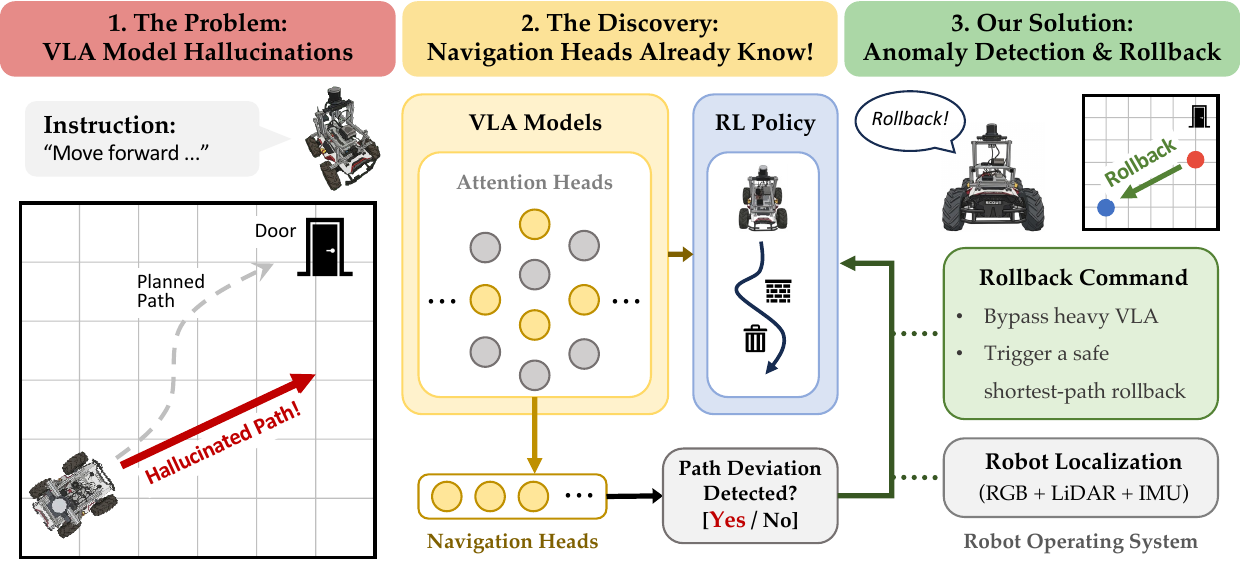}
\caption{
    Conceptual teaser of our approach to mitigating hallucinations in VLA-based navigation. 
    \textbf{1. The Problem:} VLA models are highly vulnerable to vision-language reasoning hallucinations, which cause severe deviations from planned trajectories. 
    \textbf{2. The Discovery:} We reveal that a specific subset of internal attention heads, termed Navigation Heads ($H_\texttt{nav}$), is highly sensitive to the robot's navigation states. This allows them to serve as a built-in anomaly detector with near-zero computational overhead. 
    \textbf{3. Our Solution:} Building on this insight, we enable real-time, training-free path deviation detection. Upon detecting a failure, the system immediately bypasses the computationally heavy VLA model and triggers a low-level Reinforcement Learning (RL) policy to execute a safe, collision-free rollback.
}
\vspace{-10pt}
\label{main_fig_1:teaser}
\end{figure*}

However, current path deviation detection logic primarily relies on simple heuristics like post-facto deadlocks (e.g., physical collisions or position stagnation)~\cite{hornung2014model, wellhausen2020safe, an2024etpnav}, requires expensive dedicated fine-tuning of auxiliary networks~\cite{ma2019self, ji2022proactive, taioli2024mind}, or invokes computationally heavy external LLMs via APIs~\cite{hu2023look, rajvanshi2024saynav}. Consequently, these methods fail to provide efficient, high-frequency monitoring and are inherently prohibitive for real-time deployment.

In this work, we ask a different question: Does a frozen VLA model already contain internal mechanisms that reveal whether navigation remains properly grounded? We show that the answer is yes. Specifically, we find that a small subset of attention heads consistently aligns spatiotemporal visual observations with the relevant segments of the navigation instruction. We refer to these heads as Navigation Heads ($H_\texttt{nav}$). During successful navigation, these heads exhibit a structured attention progression that tracks the agent's movement through the instruction over time. As the agent deviates, this structure degrades in characteristic ways. This observation provides an interpretable view into the model's internal navigation process and suggests that grounding failures can be detected directly from the VLA model itself, without training an auxiliary detector.

Building on this insight, we propose a training-free spatiotemporal grounding evaluation framework that formalizes both the normal evolution and the structural breakdown of $H_\texttt{nav}$. 
As illustrated in Fig.~\ref{main_fig_1:teaser}, by monitoring the attention dynamics of these heads in real time, our method produces a continuous signal of the agent's perceptual grounding state without introducing any additional trainable parameters. This enables the system to identify anomalous trajectories as they emerge and seamlessly roll back to the last verified safe state. In contrast to existing methods, our approach is directly linked to an interpretable subset of the model's internal computations, enabling effective real-time anomaly detection without incurring additional computational overhead.

Furthermore, to overcome the slow inference speeds inherent to massive VLAs~\cite{chen2024image, du2025vl, krantz2022sim, elnoor2024robot, liu2024nvila, shukor2025smolvla}, we develop a custom low-level controller based on Reinforcement Learning (RL) trained specifically for obstacle avoidance. By integrating this lightweight RL controller with the high-level VLA model, our framework ensures stable navigation under normal conditions and guarantees safe, shortest-path recovery maneuvers upon anomalous behavior detection. This hierarchical integration effectively bridges the gap between computationally heavy visual reasoning and reactive physical execution, thereby securing real-world applicability in dynamic environments. Finally, by implementing this integrated system directly on the device, we validate the practicality and robustness of our proposed architecture through extensive evaluations in both the VLN-CE virtual environment and real physical robot deployments. To the best of our knowledge, we are the first to identify navigation-specific attention heads and successfully leverage them for real-world robotic applications.

To summarize, our main contributions are as follows:
\begin{itemize}
\item We identify a small subset of attention heads in a frozen LVLM that consistently aligns spatiotemporal visual observations with linguistic navigation intent. We term these heads Navigation Heads ($H_\texttt{nav}$) and show that they expose an interpretable mechanism underlying language-guided navigation.
\item We propose a training-free anomaly detection framework that formalizes the structural evolution and breakdown of $H_\texttt{nav}$ during navigation. By monitoring attention dynamics online using only three heads, our method estimates the agent’s navigation state in real time without introducing any additional trainable parameters, achieving a 44\,\% deviation detection rate with a low false-positive rate of 10\,\%.
\item We integrate the anomaly-detection signal into a lightweight RL controller capable of reactive obstacle avoidance, thereby establishing a robust navigation-and-recovery pipeline. This integration ensures stable navigation under normal conditions and supports a shortest-path rollback behavior when an anomalous trajectory is detected.
\item We validate the proposed system in real-world robotic environments, demonstrating that our framework is highly practical beyond simulation and performs reliable embodied navigation under real deployment conditions.
\end{itemize}

%% file: tex/2_related.tex
\clearpage
\section{Related Work}
\label{sec:related}

\myparagraph{Motion-based heuristics detection.}
As a fundamental approach, this method relies on rule-based mechanisms to detect a post-facto deadlock, such as physical collisions or position stagnation. Beginning with the DWA~\cite{fox2002dynamic}, which detects entrapment in local minima, subsequent robotic systems have utilized various hardware inputs, including wheel odometry~\cite{marder2010office}, simple sensor thresholds~\cite{hornung2014model}, and multi-modal sensor fusion~\cite{wellhausen2020safe}, to reactively identify physical entrapments and execute recovery behaviors like map initialization or in-place rotations. This same logic has been recently applied to VLN. For example, ETPNav~\cite{an2024etpnav} registers a deadlock if the agent's coordinates remain unchanged after a forward action, subsequently proposing an in-place rotation heuristic.

\myparagraph{Learning-based auxiliary adapters.} To proactively detect trajectory deviations, recent paradigms rely on learning-based auxiliary adapters that necessitate training additional detection modules alongside the main navigation backbone. For instance, the Self-Monitoring Navigation Agent (SMNA)~\cite{ma2019self} introduces a progress monitor module trained with an explicit loss function to estimate goal proximity. Similarly, Speaker-Follower models~\cite{fried2018speaker} train a separate speaker network to score the alignment between candidate trajectories and instructions, while IEDL~\cite{taioli2024mind} trains a cross-modal transformer specifically to detect and localize instruction-execution errors. In physical robotics, models like PAAD~\cite{ji2022proactive} learn deep neural networks to predict future failure probabilities by combining planned motions and perception data. Other approaches, such as BADGR~\cite{kahn2021badgr} and LaND~\cite{kahn2021land}, leverage self-supervised learning or human disengagement data to train predictive models for upcoming collisions. Furthermore, methods like SCoA~\cite{zhu2021self} train a separate reinforcement learning policy to request external assistance when path uncertainty increases. Additionally, architectures such as LSTM-VAE~\cite{park2018multimodal} and LSTM encoder-decoders~\cite{malhotra2016lstm} are trained to detect anomalies by measuring reconstruction errors on multi-modal sensor streams. Consequently, these methods fundamentally require not only expensive fine-tuning of auxiliary networks, but also additional computational overhead during inference.

\myparagraph{Training-free detection via external models.} While avoiding additional fine-tuning, another line of research relies on invoking computationally heavy external Large Language Models (LLMs) or Vision-Language Models (VLMs) to reason about current states and handle path deviations in a zero-shot manner. For instance, hierarchical dual-system frameworks~\cite{hu2023look} leverage a massive VLM (System 2) to decompose instructions and guide a low-level control policy (System 1), but often suffer from latency and a lack of mutual awareness during execution errors~\cite{lin2025onetwovla}. More explicitly targeting navigation failures, frameworks such as SayNav~\cite{rajvanshi2024saynav} dynamically convert the agent's accumulated scene graphs into textual prompts, querying an external LLM to determine whether the current trajectory has deviated and to refine the plan in case of execution failures. 
However, despite being training-free, utilizing these massive external models as high-level error detectors necessitates auto-regressive text generation via APIs, which inherently introduces severe inference latency.

%% file: tex/3_method.tex
\clearpage
\section{Method}
\label{sec:method}

\begin{figure*}[!ht]
\centering
\includegraphics[width=\textwidth]{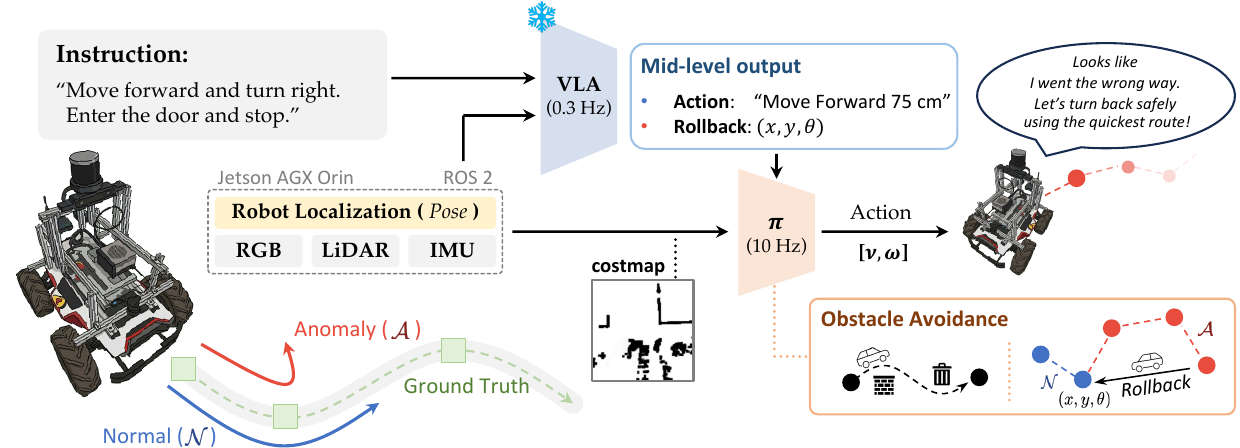}
\caption{
    \textbf{Overview of the path recovery navigation framework.} Deployed via ROS 2, our hierarchical system integrates a high-level VLA model (0.3\,Hz) and a low-level RL policy (10\,Hz). Using RGB, instructions, and pose data, the VLA detects real-time path deviations ($\mathcal{N} \to \mathcal{A}$) and triggers a rollback to the last verified checkpoint $(x, y, \theta)$. Concurrently, the RL policy ($\pi$) utilizes LiDAR costmaps to output collision-free velocity commands $[v, \omega]$. This design enables the robot to dynamically avoid obstacles and safely return to the normal path via the quickest route during recovery.
}
\vspace{-10pt}
\label{main_fig_2:overview}
\end{figure*}

\subsection{Preliminaries}
\label{ssec:pre}
\myparagraph{NaVILA Backbone.}
NaVILA~\cite{cheng2025navila} is a navigation-specific Vision-Language-Action (VLA) model built upon the VILA-8B architecture~\cite{lin2023vila}, trained via Supervised Fine-Tuning (SFT) on large-scale navigation datasets. 
To incorporate historical context, the model uniformly samples $N$ frames with a resolution of $384 \times 384$ between the initial and current frames, encoding them into $27 \times 27$ vision tokens for the LLM~\cite{grattafiori2024llama}.
These inputs are processed through $32$ layers, each with $32$ self-attention heads, allowing the model to simultaneously attend to all past frames and instruction tokens.
Based on this unified context, the model autoregressively generates mid-level action instructions (e.g., ``Turn left 30 degrees''), which are executed by a low-level controller.

\myparagraph{Distinct Characteristics of Attention Heads.}
The Transformer architecture, serving as the backbone of Large Language Models (LLMs), relies heavily on Multi-Head Self-Attention (MHSA) mechanisms. Although theoretically designed to perform parallel computations upon initialization, attention heads have been shown to spontaneously diverge during training and acquire distinct functional roles. 
Starting with the observation that specific heads attend to unique linguistic features such as syntactic relationships~\cite{clark-etal-2019-bert}, subsequent research has demonstrated that only a small subset of these specialized heads contribute to the heavy lifting~\cite{voita-etal-2019-analyzing}, allowing the majority to be pruned without significant performance degradation~\cite{michel2019sixteen}. This specialization is now recognized as a fundamental property of LLMs~\cite{zheng2024attentionheadslargelanguage} and has been shown to extend to LVLMs, where specific heads capture the semantic alignment between visual tokens and text~\cite{kang2025your}. 
Building on these insights, we analyze attention distributions to identify and leverage heads optimized for navigation tasks.

\clearpage
\begin{figure*}[!t]
\centering
\includegraphics[width=\textwidth]{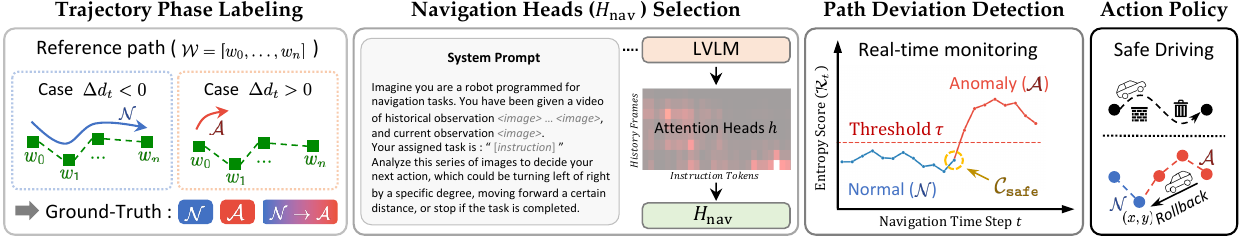}
\vspace{-15pt}
\caption{
    Our core approach consists of four consecutive stages: (i) \textbf{Phase Labeling:} Formulating ground-truth labels based on path deviation. (ii) \textbf{Head Selection:} Identifying a subset of Navigation heads $H_\texttt{nav}$ sensitive to these state transitions. (iii) \textbf{Anomaly Detection:} Monitoring the entropy ($\mathcal{R}_t$) of $H_\texttt{nav}$ to detect real-time failures and preserve safe checkpoints ($\mathcal{C}_\texttt{safe}$). (iv) \textbf{Action Policy:} Deploying an RL policy for collision-free navigation and rollbacks triggered by the detected anomalies.
}
\vspace{-15pt}
\label{main_fig_3:methods}
\end{figure*}
\subsection{Trajectory Phase Categorization}
\label{ssec:gt}

To rigorously evaluate attention heads and establish a baseline for anomaly detection, we first construct a dataset by categorizing the agent's trajectories into Normal ($\mathcal{N}$) and Anomaly ($\mathcal{A}$) phases. Given a reference path $\mathcal{W} = [w_0, \dots, w_n]$ representing a sequence of navigable waypoints for each VLN-CE episode, the navigation state evaluation is conducted through a three-step process:

(i) \textit{Target Tracking:} At each step $t$, we update the target waypoint $w^*_t$ to the closest forward waypoint along the reference path. To prevent the target from regressing, we strictly enforce a non-decreasing index constraint ($j_t \geq j_{t-1}$). 

(ii) \textit{Progress Evaluation:} We evaluate the agent's progress by calculating the distance delta $\Delta d_t$. If the agent reaches a new waypoint, we set $\Delta d_t = 0$. If the target remains unchanged, $\Delta d_t$ is the change in distance to the current target ($d_t - d_{t-1}$, where $d_t$ is the Euclidean distance to $w^*_t$). Consequently, a positive $\Delta d_t$ intuitively indicates a path deviation. 

(iii) \textit{Phase Classification:} We employ a one-way state machine with patience $p$. Starting in the $\mathcal{N}$ phase, if $p$ consecutive translational steps yield $\Delta d_t > 0$, the state irreversibly transitions to $\mathcal{A}$, applying labels retroactively from the first deviation step. Rotational actions are excluded in $\mathcal{N}$ but included in $\mathcal{A}$ as continued disorientation. Crucially, if $p$ consecutive on-track steps occur during $\mathcal{A}$, the episode is truncated to exclude noisy corrections. 

Consequently, episodes are categorized into only normal ($\mathcal{N}$), only anomaly ($\mathcal{A}$), and normal to anomaly ($\mathcal{N} \to \mathcal{A}$), as shown in Fig.~\ref{main_fig_3:methods}.

\subsection{Navigation Heads Selection}
\label{ssec:nav_heads}

\begin{figure*}[!t] 
\centering
\includegraphics[width=\textwidth]{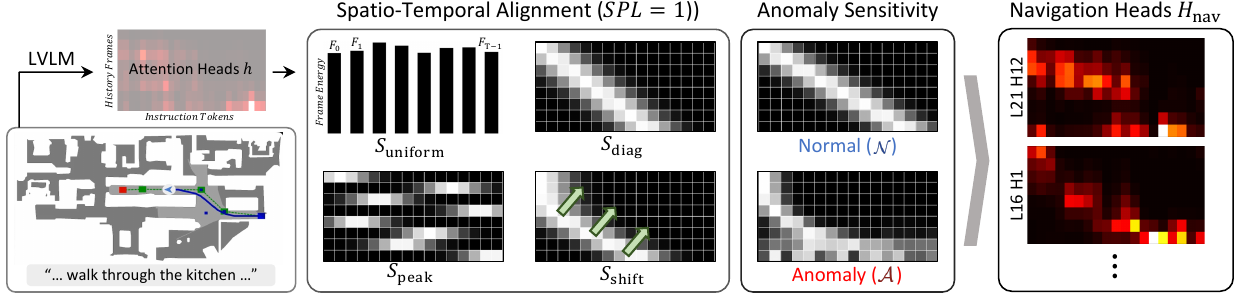}
\vspace{-15pt}
\caption{Training-free spatiotemporal grounding framework for navigation.
To select robust navigation heads, candidates are first evaluated using the alignment score $I_{\texttt{diag}}(h)$, which comprises even frame energy ($S_{\texttt{uniform}}$), focused instruction attention ($S_{\texttt{peak}}$), diagonal alignment ($S_{\texttt{diag}}$), and smooth transition ($S_{\texttt{shift}}$), and subsequently assessed for their sensitivity to attention changes between normal and anomalous navigation.
}
\vspace{-10pt}
\label{main_fig_4:navigation_head}
\end{figure*}

Leveraging the distinct functional roles of LVLM attention heads, we identify a navigation-critical subset ($H_\texttt{nav}$). This subset is defined by two core criteria: (i) explicit spatio-temporal alignment, and (ii) high sensitivity to the cognitive anomalies defined in Sec.~\ref{ssec:gt}.

\myparagraph{Spatio-Temporal Alignment.} 
We hypothesize that specific attention heads $h$ capture the temporal causality between the visual sequence of $T$ history frames $\mathcal{V} = \{F_k\}_{k=0}^{T-1}$ ($F_0$ is the oldest observation, $F_{T-1}$ the current frame) and the sequential navigation instruction. To isolate these heads, we extract the Instruction-to-Frame attention weights to form a 2D alignment matrix $A \in \mathbb{R}^{T \times N}$ for $N$ instruction tokens. We score these heads using a novel Spatio-Temporal alignment score $I_{\texttt{diag}}(h)$ that evaluates observation-level grounding and dynamic spatiotemporal shifts based on base attention dynamics.

First, observation-level cross-modal grounding is validated via frame energy uniformity ($S_\texttt{uniform}$) and focus sharpness ($S_\texttt{peak}$). To ensure that the head continuously grounds visual observations without ignoring any frame in the temporal sequence, $S_\texttt{uniform}$ penalizes vanishing attention mass by measuring the ratio of each frame's energy to the peak episode energy. Concurrently, to verify that the agent precisely targets a specific instruction segment rather than dispersing its focus, $S_\texttt{peak}$ ensures a sharp, unimodal peak. 
Given the attention weight $A_{k,j}$ from frame $k$ to token $j$, we formulate the frame energy $E_k = \sum_{j=0}^{N-1} A_{k,j}$, the normalized distribution $P_{k,j} = A_{k,j} / E_k$, and the temporal center of mass $c_k = \sum_{j=0}^{N-1} j \cdot P_{k,j}$. Defining $\sigma_k$ and $m_k$ as the standard deviation and local window mass of $P_k$ around $c_k$, respectively, we compute:
\begin{align}\label{method_eq:1}
    S_\texttt{uniform} = \frac{1}{T} \sum_{k=0}^{T-1} \frac{E_k}{\max_{k'} E_{k'}}, \quad S_\texttt{peak} = \frac{1}{T} \sum_{k=0}^{T-1} \frac{1}{2} \left( \left( 1 - \frac{\sigma_k}{\sigma_\text{max}} \right) + m_k \right),
\end{align}
where $k' \in \{0, \dots, T-1\}$ indexes the frames for maximum energy, and $\sigma_\text{max}$ represents the standard deviation of a uniform distribution over instruction length.

Second, to evaluate the dynamic spatiotemporal shifting trajectory, we define the ideal attention peak $i^{*}_k$ for each frame, which forms a diagonal across the 2D matrix $A$. Accordingly, we compute the spatial proximity to this ideal diagonal ($S_\texttt{diag}$) and penalize erratic jumps to enforce a smooth shift towards the latter part of the instruction sequence as the navigation progresses ($S_\texttt{shift}$):

\begin{align}\label{method_eq:2}
    & S_\texttt{diag} = 1 - \frac{1}{T} \sum_{k=0}^{T-1} \frac{|c_k - i^{*}_k|}{N-1}, \quad \text{where } i^{*}_k = \left( \frac{k}{T-1} \right) \cdot (N-1), \nonumber \\
    & S_\texttt{shift} = \frac{1}{2(T-1)} \sum_{k=1}^{T-1} \left( \mathbb{I}(\Delta c_k > 0) + \exp\left( -\frac{1}{2} \left( \frac{\Delta c_k}{\Delta i^{*}_k} - 1 \right)^2 \right) \right),
\end{align}
where $\Delta c_k = c_k - c_{k-1}$ and $\Delta i^{*}_k = i^{*}_k - i^{*}_{k-1}$, with $\mathbb{I}(\cdot)$ as the indicator function.

Ultimately, to establish a robust baseline, the alignment score $I_{\texttt{diag}}(h)$ is calculated as the expectation over $\mathcal{E}_\text{ideal}$, a set of optimal episodes achieving a perfect Success weighted by Path Length (SPL)~\cite{anderson2018evaluation} of 1.0:

\begin{align}\label{method_eq:3}
    I_{\texttt{diag}}(h) = \mathbb{E}_{\tau \sim \mathcal{E}_\text{ideal}} \left[ S_\texttt{uniform}^{(h)} \cdot S_\texttt{peak}^{(h)} \cdot \left( \lambda S_{\texttt{diag}}^{(h)} + (1 - \lambda) S_{\texttt{shift}}^{(h)} \right) \right],
\end{align}
where $\lambda$ balances spatial precision with temporal smoothness. 
Heads consistently yielding high $I_{\texttt{diag}}(h)$ are selected as initial candidates for $H_\texttt{nav}$.

\myparagraph{Cognitive Anomaly Sensitivity.}
While the alignment score $I_{\texttt{diag}}$ identifies heads optimized for instruction following, a robust navigation head must also exhibit distinct attentional shifts during failure modes. Therefore, we evaluate the sensitivity of candidate heads by analyzing the divergence in their focus sharpness ($S_\texttt{peak}$) distributions between the normal ($\mathcal{N}$) and anomalous ($\mathcal{A}$) phases. This degree of separation for each head $h$ is quantified using Cohen's $d$~\cite{lakens2013calculating}:

\begin{align}\label{method_eq:cohen}
    d(h) = \frac{|\mu_{\mathcal{N}} - \mu_{\mathcal{A}}|}{\sigma_{\text{pooled}}}, \quad \text{where } \sigma_{\text{pooled}} = \sqrt{\frac{(n_{\mathcal{N}}-1)\sigma_{\mathcal{N}}^2 + (n_{\mathcal{A}}-1)\sigma_{\mathcal{A}}^2}{n_{\mathcal{N}}+n_{\mathcal{A}}-2}}
\end{align}
Here, $\mu$ and $\sigma$ represent the mean and standard deviation of the focus sharpness within each partition, with $n$ denoting the sample count.
A higher $d(h)$ signifies a substantial shift or collapse in attention focus when the agent becomes stuck. Consequently, we rank the candidate heads by $d(h)$ to form $H_\texttt{nav}$.

\subsection{Attention-Based Path Deviation Detection}
\label{ssec:anomaly}

To detect path deviations, we compute the mean attention entropy $E_t$ at each time step $t$ using the top-$K$ heads from $H_\texttt{nav}$. We then monitor the relative entropy score $\mathcal{R}_t$, as the ratio of $E_t$ to its rolling window average of size $W$:
\begin{equation}\label{method_eq:reldiff}
    \mathcal{R}_t = \frac{E_t}{\frac{1}{W} \sum_{i=1}^{W} E_{t-i} + \epsilon}, \quad \text{where } E_t = \frac{1}{|H_\texttt{nav}|} \sum_{h \in H_\texttt{nav}} \mathcal{H}(A_t^{(h)}).
\end{equation}
Here, $\mathcal{H}(\cdot)$ denotes the normalized Shannon entropy averaged over visual frames, ensuring scale invariance. A high $\mathcal{R}_t$ indicates a sudden dispersion of attention relative to the agent's recent history, serving as a strong precursor to navigation failure.
Mirroring the patience mechanism of the ground-truth labeling, an anomaly state ($\mathcal{A}$) is registered only if $\mathcal{R}_t$ exceeds a natural threshold $\tau$ for $P$ consecutive steps. Throughout the normal phase ($\mathcal{N}$), we continuously update a safe checkpoint $\mathcal{C}_\text{safe}=\{s_t, \mathcal{V}_t, \mathcal{M}_t, A_t\}$, preserving the agent's pose $s_t=(x, y, z, \theta)$, visual history $\mathcal{V}_t$, the entropy buffer $\mathcal{M}_t = \{E_{t-W}, \dots, E_{t-1}\}$, and the attention state $A_t$ required for future recovery. The hyperparameters $(K, W, \tau, P)$ are jointly optimized via a grid sweep on the training split, as detailed in Sec.~\ref{sec:exp}.

\clearpage
\subsection{Collision-Free Low Level Action Policy}
\label{ssec:rl}

\begin{figure*}[!ht]
\centering
\includegraphics[width=\textwidth]{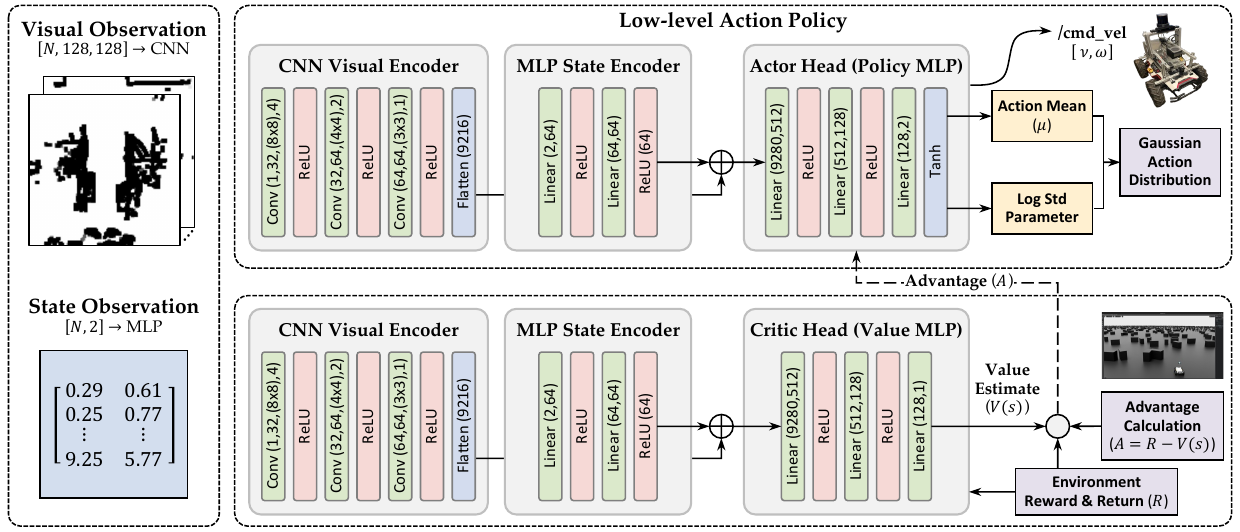}
\caption{
    The network processes $128\times128$ visual costmaps through a CNN encoder to extract spatial obstacle features, and relative local subgoal states $(x,y)$ via an MLP encoder for directional guidance.
    These features are fused to feed two specialized heads: the \textbf{Actor head}, which outputs Gaussian distributions for velocity control commands $(\upsilon,\omega)$, and the \textbf{Critic head}, which estimates the state-value $V(s)$. This hierarchical design enables the agent to learn reactive obstacle avoidance and robust subgoal navigation through advantage-based reinforcement learning.
}
\vspace{10pt}
\label{main_fig_5:rl_architecture}
\end{figure*}

To achieve safe collision avoidance and low-level control, we mirror a customized real-world robotic platform and its sensor suite within the physics-based simulator IsaacLab~\cite{mittal2025isaaclab}. The agent is trained using Proximal Policy Optimization (PPO)~\cite{schulman2017proximal} with relative goal coordinates and LiDAR data.
To ensure the policy remains agnostic to specific LiDAR hardware configurations, we employ a 2D occupancy grid costmap as an intermediate observation space rather than using raw point clouds. Specifically, we apply ego-motion compensation to spatially align the map with the robot's motion, and we couple it with a temporal decay mechanism that gradually fades historical obstacle data. This formulation enables the system to robustly handle dynamic obstacles and mitigate the impact of temporary sensor blind spots.
To process these inputs, we design a lightweight and efficient actor-critic architecture (as shown in Fig.~\ref{main_fig_5:rl_architecture}) that decouples the observation streams.
A CNN extracts spatial features from the costmap for obstacle avoidance, while an MLP encodes the relative direction and distance to the goal for path tracking. These feature representations are subsequently fused to predict control actions. 
Furthermore, to promote generalization, the policy is trained in a large-scale 250\,m $\times$ 200\,m virtual environment populated with 2,000 procedural obstacles (cylinders and boxes), with a minimum traversable clearance of 3.0\,m strictly enforced.
We formulate a 14-term dense reward to jointly optimize goal-reaching, safety, and smoothness. A curriculum learning strategy progressively extends the goal distance from 5\,m to 20\,m, ensuring reliable long-horizon evasive maneuvers by consistently accumulating goal-reaching rewards. Finally, training is heavily accelerated via CUDA batching. Detailed hyperparameters are provided in the Supplementary Material.

\clearpage
\subsection{Sim-to-Real Transfer Environment.}
\label{ssec:software}

To bridge the sim-to-real gap, we designed an optimized on-device architecture within the Robot Operating System (ROS)~\cite{macenski2022robot} that supports both the VLA model and the RL policy. Both strictly require real-time state estimation: the VLA relies on localization for failure recovery, while the RL policy requires precise odometry for costmap ego-motion compensation.
To mitigate computational bottlenecks, we adopted Fast-LIVO2~\cite{zheng2024fast}, a lightweight, CPU-based multi-sensor odometry framework. To establish a fixed global coordinate system, we fused this odometry with AprilTag~\cite{olson2011apriltag} detections via an Extended Kalman Filter (EKF). This approach enables reliable data comparison and ensures that limited GPU resources are reserved exclusively for the computationally intensive AI models, thereby maximizing real-time inference efficiency. Comparative experiments are detailed in the Supplementary Material.
Regarding robot locomotion, the raw velocity commands $(v,w)$ generated by the RL policy often exhibit high-frequency oscillations that compromise physical stability. To ensure safe deployment, we implemented a kinematic action smoother utilizing a low-pass filter, acceleration clipping, and deadzone masking.
By enforcing a strictly bounded acceleration profile, the smoother preserves the agent's reactive obstacle avoidance capabilities while mitigating mechanical strain and preventing motor overload. Detailed implementation specifics are provided in the Supplementary Material.

%% file: tex/4_experiment.tex
\section{Experiments}
\label{sec:exp}

\subsection{Experimental Setup}
\myparagraph{Simulation and Training.} 
Experiments and RL training were conducted on a single NVIDIA RTX 6000 Ada GPU. We adopt the default NaVILA~\cite{cheng2025navila} configuration ($384 \times 384$ resolution), excluding the detection-specific hyperparameters and ablation studies detailed in Sec.~\ref{anomaly}. Further details regarding the Isaac Lab environment are provided in the Supplementary Material.

\myparagraph{Real-world Deployment.}
The framework was deployed on an AgileX Scout 2.0 mobile robot equipped with an NVIDIA Jetson AGX Orin (64\,GB), a ZED 2i camera, and an RS-LiDAR. Remote control is managed via a host PC over 5G, following the setup in~\cite{jeong2025agrichrono}. Detailed communication protocols and calibration parameters are available in the Supplementary Material.

\subsection{Evaluating Navigation Heads for Anomaly Detection}
\label{anomaly}

\begin{table}[!ht]
\caption{
    Ground Truth (GT) analysis of trajectory phases (Sec.~\ref{ssec:gt}). On NaVILA~\cite{cheng2025navila} trajectories in VLN-CE R2R~\cite{anderson2018vision}, high Cohen's $d$ ($>1.2$) confirms strong statistical separation in waypoint distances between $\mathcal{N}$ and $\mathcal{A}$ steps. Notably, the Val-Unseen `Only $\mathcal{N}$' ratio (56.0\,\%) matches the baseline Success Rate, verifying GT reliability.
}
\vspace{-5pt}
\centering
\renewcommand{\arraystretch}{1.1}
\setlength\tabcolsep{2.4pt}
\small
\adjustbox{max width=\columnwidth}{
    \begin{tabular}{c|c|c|c|c|c|c|c}
        \hlineB{2.5}
        \multirow{2}{*}{\textbf{Split}} & \multicolumn{4}{c|}{\textbf{Episode Category}} & \multicolumn{3}{c}{\textbf{Path Distance \& Effect Size}} \\
        \cline{2-8}
         & \textbf{Total} & \textbf{Only $\mathcal{N}$} & \textbf{Only $\mathcal{A}$} & \textbf{$\mathcal{N} \to \mathcal{A}$} & \textbf{$\mathcal{N}$ Median} & \textbf{$\mathcal{A}$ Median} & \textbf{Cohen's $d$} \\
        \hline
        Train & 1,000 & 617 (61.7\,\%) & 79 (7.9\,\%) & 304 (30.4\,\%) & 0.13\,m & 2.38\,m & 1.54 \\
        Val-Seen & 778 & 435 (55.9\,\%) & 91 (11.7\,\%) & 252 (32.4\,\%) & 0.14\,m & 2.97\,m & 1.60 \\
        Val-Unseen & 1,839 & 1,029 (56.0\,\%) & 263 (14.3\,\%) & 547 (29.7\,\%) & 0.16\,m & 2.53\,m & 1.27 \\
        \hlineB{2.5}
    \end{tabular}
}
\vspace{-17pt}
\label{tab:gt_validation}
\end{table}

\myparagraph{Validation of Ground Truth Labels.}
As summarized in Table~\ref{tab:gt_validation}, we rigorously validate our behavioral Ground Truth (GT) labels to ensure the reliability of the anomaly detection framework. For this validation, we utilize the NaVILA~\cite{cheng2025navila} backbone to extract trajectories from 1,000 random episodes in the Train split, as well as all episodes in the Val-Seen and Val-Unseen splits of the VLN-CE R2R dataset~\cite{anderson2018vision}. Each episode is categorized into mutually exclusive classes.
Step-level path-distance analysis reveals that, even without explicit spatial thresholds, our state-machine-based labeling successfully captures distinct navigational behaviors. Specifically, $\mathcal{N}$ steps tightly adhere to the optimal path with median deviations of $0.13$--$0.16\,\text{m}$, while $\mathcal{A}$ steps diverge significantly to $2.38$--$2.97\,\text{m}$. The resulting large effect size (Cohen's $d > 1.2$) statistically confirms that our behavior-driven criteria robustly isolate severe deviations from minor steering noise. Furthermore, the only $\mathcal{N}$ ratio in the Val-Unseen split ($56.0\,\%$) matches the baseline's terminal Success Rate, justifying the macroscopic validity of our GT.
Finally, analyzing the temporal structure of $\mathcal{N} \to \mathcal{A}$ transitions reveals a median $\mathcal{N}$ phase of $6$--$7$ steps followed by an $\mathcal{A}$ phase of $9$--$10$ steps. This balanced phase distribution confirms that the anomaly-onset positions are assigned, providing sufficient baseline history to characterize the normal navigational context and an adequate temporal window for detection.

\begin{figure}[!b]
\vspace{-20pt}
\centering
    \begin{subfigure}{0.33\linewidth}
        \includegraphics[width=\linewidth, trim=10pt 0pt 10pt 0pt, clip]{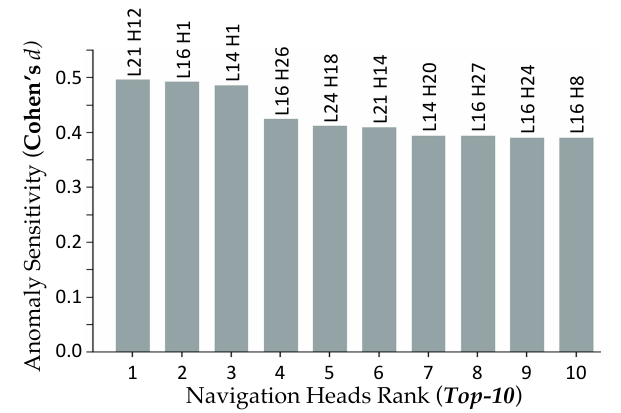}
        \caption{$H_\texttt{nav}$ Discriminability}
        \label{fig:sub_fig6}
    \end{subfigure}
    \hspace{-5pt}
    \begin{subfigure}{0.33\linewidth}
        \includegraphics[width=\linewidth, trim=10pt 0pt 10pt 0pt, clip]{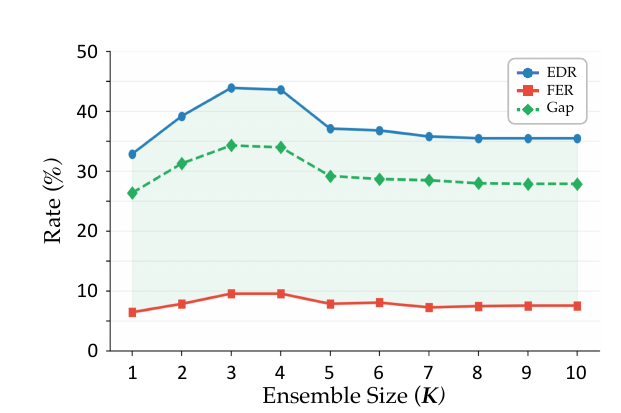}
        \caption{Episode-level
        \textit{K} ablation}
        \label{fig:sub_fig7}
    \end{subfigure}
    \hfill
    \begin{subfigure}{0.33\linewidth}
        \includegraphics[width=\linewidth, trim=10pt 0pt 10pt 0pt, clip]{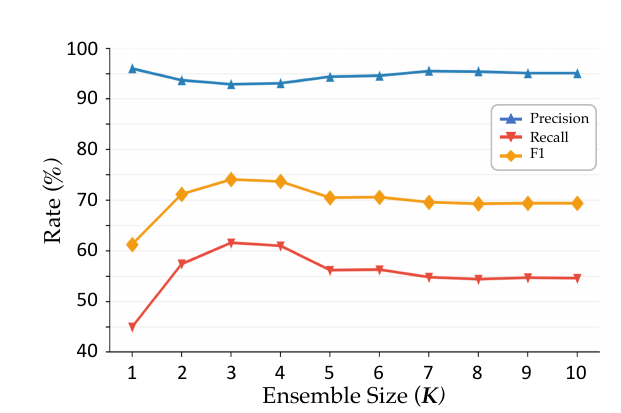}
        \caption{Step-level \textit{K} ablation}
        \label{fig:sub_fig8}
    \end{subfigure}
    \vspace{-3pt}
    \caption{
        Ablation study of navigation heads ($H_\texttt{nav}$) and ensemble size ($K$) on the R2R train split. 
        (a) Discriminability of attention heads measured by Cohen's $d$; higher values indicate greater sensitivity to anomalous behaviors. 
        (b) Episode-level detection performance across $K$. Thresholds are optimized via grid search to maximize Episode Detection Rate (EDR) while constraining False Episode Rate (FER) within 5--10\,\% (Gap = EDR $-$ FER). 
        (c) Step-level classification performance on $\mathcal{N} \to \mathcal{A}$ episodes. 
        Notably, an ensemble of just $K=3$ heads achieves 43.9\,\% EDR and 9.6\,\% FER at the episode level, alongside 92.9\,\% Precision, 61.6\,\% Recall, and 74.1\,\% F1 at the step level.
    }
\vspace{-5pt}
\label{fig:heads_ensemble}
\end{figure}

\begin{table}[!ht]
\caption{
    Performance comparison of training-free anomaly detection on R2R validation splits. 
    Evaluating $H_\texttt{nav}$ and hyperparameters derived solely from the train split, we baseline against rule-based heuristics~\cite{marder2010office, an2024etpnav} to ensure a fair comparison with near-zero inference overhead. 
    Our approach significantly outperforms these baselines in both episode-level net advantage (Gap) and step-level F1. 
    Notably, while heuristics suffer from severe recall bottlenecks (missing >60\,\% of anomalies), our method robustly captures navigation failures without requiring additional trainable modules.
}
\vspace{-5pt}
\centering
\renewcommand{\arraystretch}{1.4}
\setlength\tabcolsep{2.4pt}
\small
\adjustbox{max width=\columnwidth}{
    \begin{tabular}{l|ccc|ccc|ccc|ccc}
        \hlineB{2.5}
        \multirow{3}{*}{\textbf{Method}} & \multicolumn{6}{c|}{\textbf{R2R Val-Seen}} & \multicolumn{6}{c}{\textbf{R2R Val-Unseen}} \\
        \cline{2-13}
         & \multicolumn{3}{c|}{\textbf{All Episodes}} & \multicolumn{3}{c|}{\textbf{$\mathcal{N} \to \mathcal{A}$ Step-level}} & \multicolumn{3}{c|}{\textbf{All Episodes}} & \multicolumn{3}{c}{\textbf{$\mathcal{N} \to \mathcal{A}$ Step-level}} \\
        \cline{2-13}
         & \textbf{EDR} $\uparrow$ & \textbf{FER} $\downarrow$ & \textbf{Gap} $\uparrow$ & \textbf{Prec.} $\uparrow$ & \textbf{Rec.} $\uparrow$ & \textbf{F1} $\uparrow$ & \textbf{EDR} $\uparrow$ & \textbf{FER} $\downarrow$ & \textbf{Gap} $\uparrow$ & \textbf{Prec.} $\uparrow$ & \textbf{Rec.} $\uparrow$ & \textbf{F1} $\uparrow$ \\
        \hline
        Stagnation~\cite{marder2010office} & 24.5\,\% & \textbf{4.8\,\%} & 19.7\,\% & \textbf{99.7\,\%} & 31.3\,\% & 47.7\,\% & 29.6\,\% & \textbf{5.9\,\%} & 23.7\,\% & \textbf{95.6\,\%} & 38.7\,\% & 55.1\,\% \\
        Act.Failure~\cite{an2024etpnav} & 0.3\,\% & 6.0\,\% & $-$5.7\,\% & 70.6\,\% & 2.8\,\% & 5.5\,\% & 1.5\,\% & 6.3\,\% & $-$4.8\,\% & 77.9\,\% & 5.6\,\% & 10.4\,\% \\
        \hline
        \textbf{Ours} & \textbf{44.6\,\%} & 11.7\,\% & \textbf{32.9\,\%} & 91.3\,\% & \textbf{68.6\,\%} & \textbf{78.3\,\%} & \textbf{41.9\,\%} & 9.6\,\% & \textbf{32.2\,\%} & 92.5\,\% & \textbf{65.1\,\%} & \textbf{76.4\,\%} \\
        \hlineB{2.5}
    \end{tabular}
}
\vspace{-10pt}
\label{tab:baseline_comparison}
\end{table}

\begin{figure}[!b]
\vspace{-20pt}
\centering
    \begin{subfigure}{0.33\linewidth}
        \includegraphics[width=\linewidth, trim=8pt 0pt 8pt 0pt, clip]{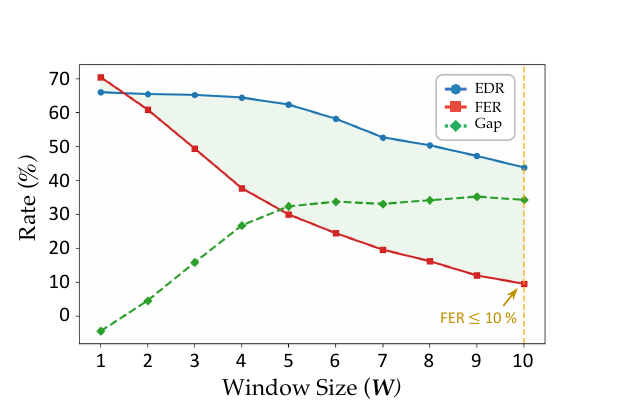}
        \caption{Effect of Window size}
        \label{fig:sub_fig9}
    \end{subfigure}
    \hspace{-10pt}
    \begin{subfigure}{0.33\linewidth}
        \includegraphics[width=\linewidth, trim=8pt 0pt 8pt 0pt, clip]{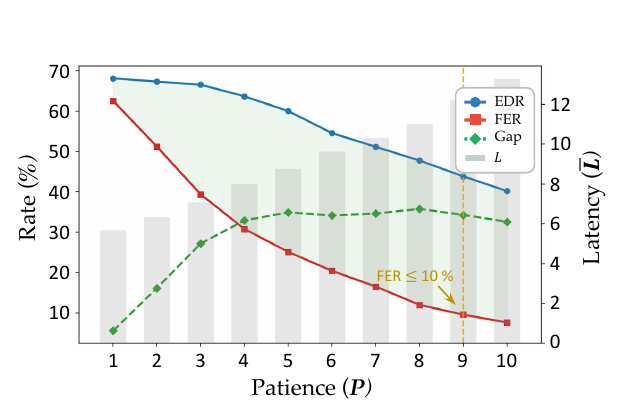}
        \caption{Effect of Patience}
        \label{fig:sub_fig10}
    \end{subfigure}
    \hspace{5pt}
    \begin{subfigure}{0.33\linewidth}
        \includegraphics[width=\linewidth, trim=8pt 0pt 8pt 0pt, clip]{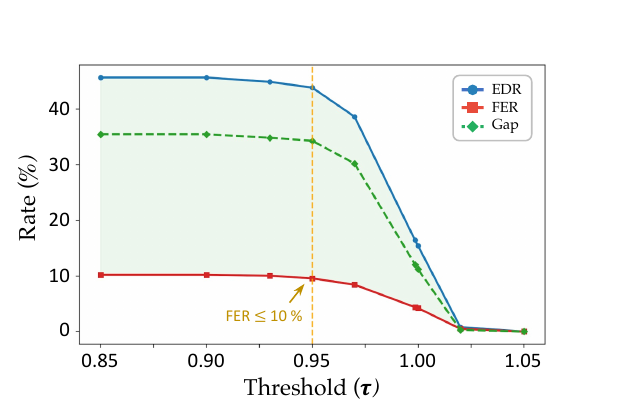}
        \caption{Effect of Threshold}
        \label{fig:sub_fig11}
    \end{subfigure}
    \caption{
        Hyperparameter grid search for the proposed anomaly detector ($K=3$) on the R2R train split. 
        (a) Both EDR and detection Gap (EDR $-$ FER) increase monotonically, saturating around $W=7$. 
        (b) Increasing $P$ effectively suppresses FER, albeit at the cost of a linearly increasing mean detection latency. 
        (c) The detector exhibits sharp sensitivity near $\tau=1.0$, where both EDR and FER drop rapidly. 
        Ultimately, driven by the FER $< 10\,\%$ constraint, we select ($W=10, P=9, \tau=0.95$) that achieves an EDR of 43.9\,\%, thereby preventing unnecessary interruptions during normal navigation.
    }
\vspace{-10pt}
\label{fig:effect}
\end{figure}

\myparagraph{Path Deviation Detection Performance.}
To verify the generalization of our methodology, we first derive the optimal configuration for our anomaly detector strictly from the R2R train split. As shown in Fig.~\ref{fig:sub_fig6}, we extract a subset of attention heads ($H_\texttt{nav}$) and conduct an extensive grid search over 9,000 combinations, comprising the ensemble size ($K \in \{1, \dots, 10\}$), patience ($P \in \{1, \dots, 10\}$), sliding window ($W \in \{1, \dots, 10\}$), and 9 values for the activation threshold ($\tau$). 
As illustrated in Fig.~\ref{fig:sub_fig7}, by strictly constraining the False Episode Rate (FER)---the rate at which normal navigation is misclassified as anomalous---below 10\,\%, the ensemble combination of $K=3$ achieves the highest Episode Detection Rate (EDR), defined as the rate of correctly identifying anomalous episodes. 
At the step level, evaluated across 304 $\mathcal{N} \to \mathcal{A}$ transition scenarios (consisting of 2,277 normal and 4,271 anomalous steps), this configuration records 92.9\,\% Precision, 61.6\,\% Recall, and a 74.1\,\% F1 score. The individual performance trends concerning $P$, $W$, and $\tau$ are detailed in Fig.~\ref{fig:effect}.

\begin{figure*}[!ht]
\centering
\includegraphics[width=\textwidth]{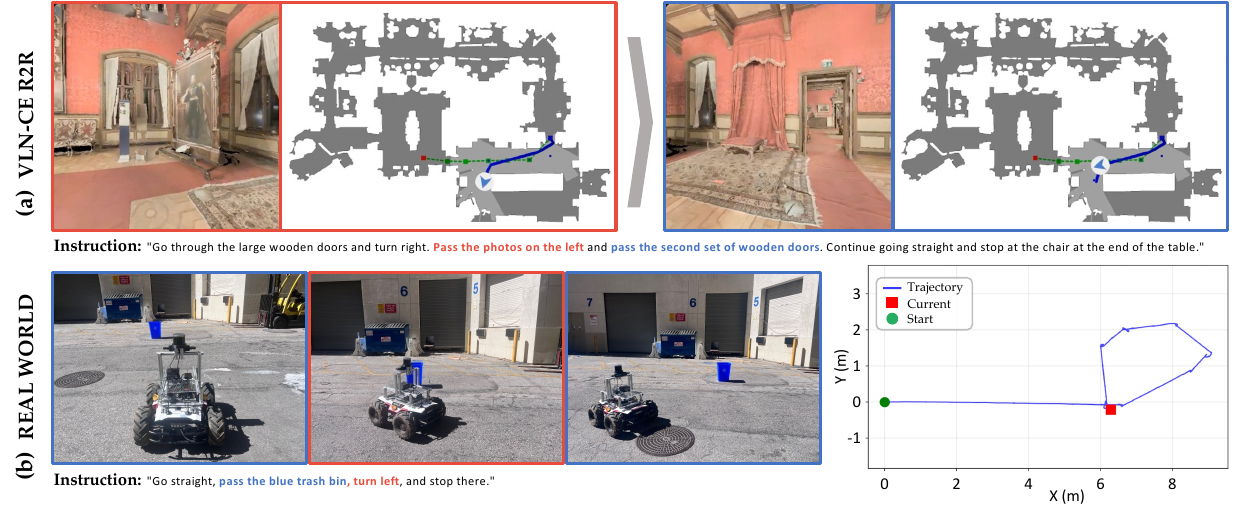}
\vspace{-20pt}
\caption{
    (a) Hallucination-induced path deviation and recovery in VLN-CE simulation.
    (b) Real-robot visualization of RL-driven recovery from path deviations during VLA navigation.
    Additional results are provided in the Supplementary Material.
}
\vspace{-10pt}
\label{main_fig_12:qualitative}
\end{figure*}

\noindent Table~\ref{tab:baseline_comparison} compares our approach against established zero-inference-overhead heuristics~\cite{marder2010office, an2024etpnav} on the validation splits, strictly utilizing the heads and hyperparameters derived from the train data. Our method significantly outperforms these rigid baselines across both episode- and step-level metrics. Position Stagnation~\cite{marder2010office}, which flags anomalies via fixed $(x,y)$ displacement thresholds, maintains a low FER (4.8\,\% on Val-Seen) but suffers from severe recall bottlenecks (missing >60\,\% of step anomalies), yielding a maximum Gap of only 23.7\,\% on Val-Unseen. Action Failure~\cite{an2024etpnav}, which monitors coordinate shifts strictly during forward actions, similarly fails to reliably detect deviations, resulting in a negative Gap. In contrast, our method robustly captures navigation failures by monitoring the semantic misalignment between linguistic instructions and visual observations, rather than relying solely on the robot's physical kinematics. Consequently, on the Val-Unseen split, our detector achieves a Gap of 32.2\,\% and a step-level F1 score of 76.4\,\%, driven by a substantial improvement in recall (65.1\,\%). This confirms that leveraging intrinsic cross-modal attention dynamics yields a far more comprehensive and generalizable signal for path deviation than coordinate-based mechanisms. Qualitative results are in Fig.~\ref{main_fig_12:qualitative}

\subsection{RL Obstacle Avoidance Performance}

\begin{figure}[!ht]
\centering
    \begin{subfigure}{0.4\linewidth}
        \centering        
        \small
        \renewcommand{\arraystretch}{1.2}
        \adjustbox{max width=\linewidth}{
        \setlength\tabcolsep{2.4pt}        
        \begin{tabular}{c|c|c|c}
            \hlineB{2.5}
            \textbf{Metric} & \textbf{SR} $\uparrow$ & \textbf{CR} $\downarrow$ & \textbf{TR} $\downarrow$ \\
            \hline
            APF~\cite{khatib1986real} & 57.4\,\% & 7.8\,\% & 34.5\,\% \\
            DWA~\cite{fox2002dynamic} & 24.1\,\% & 73.1\,\% & 0.7\,\% \\
            MPPI~\cite{williams2017information} & 3.4\,\% & 86.1\,\% &0.4\,\%  \\
            TEB~\cite{rosmann2012trajectory} & 55.4\,\% & 37.5\,\% &6.9\,\%  \\
            \hline
            \textbf{Ours} & \textbf{87.3\,\%} & \textbf{10.6\,\%} & \textbf{2.0\,\%} \\
            \hlineB{2.5}
        \end{tabular}
        }
        \caption{RL Policy Performance}
        \label{tab:rl_metric}
    \end{subfigure}
    \hfill
    \begin{subfigure}{0.57\linewidth}
        \includegraphics[width=\linewidth]{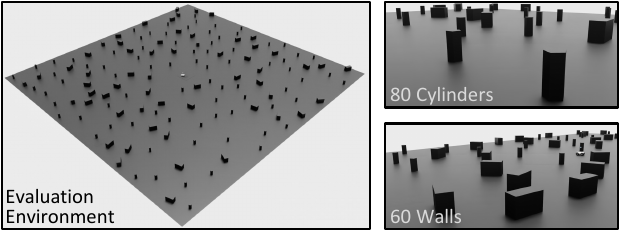}
        \caption{Evaluation Environments}
    \label{fig:rl_visualization}
    \end{subfigure}
    \vspace{-5pt}
    \caption{
        \textbf{Evaluation of the low-level RL obstacle avoidance policy.} 
        (a) Quantitative comparison of average performance metrics (SR, CR, and TR) against classical local planning baselines. Our proposed policy significantly outperforms traditional methods, achieving the highest average success rate of 87.3\,\%. 
        (b) Visualization of the IsaacLab evaluation environment. The testbed comprises a $60\,\text{m} \times 60\,\text{m}$ subterrain populated with procedurally generated cylindrical obstacles and wall-shaped boxes to comprehensively assess collision-free navigation.
    }
\end{figure}

\myparagraph{Evaluation Setup.}
We evaluate our RL obstacle avoidance policy in IsaacLab~\cite{mittal2025isaac}. The testbed comprises a single $60\,\text{m} \times 60\,\text{m}$ subterrain (Fig.~\ref{fig:rl_visualization}) populated with 80 cylindrical obstacles and 60 wall-shaped boxes (all $2.0\,\text{m}$ high). A minimum clearance of $2.0\,\text{m}$ ensures the existence of feasible paths. Across four target-distance scenarios ($5\,\text{m}$ to $20\,\text{m}$), 512 parallel agents are spawned, yielding 2,048 independent episodes. The policy maps local costmaps to continuous velocity commands bounded to $v \in [-0.5, 0.5]\,\text{m/s}$ and $\omega \in [-1.0, 1.0]\,\text{rad/s}$. Episodes conclude upon: (i) reaching within $0.5\,\text{m}$ of the target (Success Rate, SR); (ii) stalling $>3.0\,\text{s}$ or exceeding $150.0\,\text{s}$ (Terminate Rate, TR); or (iii) colliding with an obstacle exceeding $200\,\text{N}$ of force (Collision Rate, CR).

\myparagraph{Evaluation Analysis.}
Table~\ref{tab:rl_metric} compares our policy against four classical local planners using purely local observations (a $128 \times 128$ costmap and relative goal position). Among the baselines, APF~\cite{khatib1986real} attains a relatively high SR (57.4\,\%) but suffers a severe TR (34.5\,\%) due to frequent entrapment in local minima. TEB~\cite{rosmann2012trajectory} shows a comparable SR (55.4\,\%) but struggles with collisions (CR 37.5\,\%) in dense configurations. DWA~\cite{fox2002dynamic} and MPPI~\cite{williams2017information} fail entirely (CR $>73\,\%$), as their short-horizon trajectory sampling cannot anticipate complex obstacle layouts. In contrast, our RL policy achieves the highest SR (87.3\,\%), outperforming the best baseline by nearly 30 percentage points, while maintaining a low CR (10.6\,\%) and TR (2.0\,\%). This demonstrates that our tailored reward function and extensive environmental interaction effectively equip the policy with implicit spatial reasoning, allowing it to smoothly navigate dense layouts where classical reactive planners crash or stall.
Detailed per-distance metrics are provided in the Supplementary Material.

\clearpage

\subsection{Real-world Evaluation}
As shown in Fig.~\ref{main_fig_12:qualitative}, we evaluated our framework on a custom VLA-driven robot. Upon detecting deviations, an RL policy safely executes a trajectory rollback. Setup details and further experiments are in the Supplementary Material.

%% file: tex/5_conclusion.tex
\section{Conclusion}
\label{con}

We present a training-free anomaly detection and recovery framework for VLA models. By analyzing the internals of a frozen VLA, we discovered a sparse set of Navigation Heads ($H_\texttt{nav}$) that naturally capture the spatiotemporal alignment between visual observations and linguistic instructions. Building on this, we propose a real-time system that detects path deviations by measuring the entropy of these specific heads without additional computational overhead. To ensure safe physical execution, we integrated this detection signal with a lightweight, low-level RL policy capable of reactive obstacle avoidance and shortest-path rollback. Evaluations in both simulations and real-robot experiments demonstrate that our approach effectively mitigates navigation failures caused by VLA hallucinations. Ultimately, this research provides a practical solution for deploying reliable, self-monitoring embodied agents in complex continuous environments.

\myparagraph{Limitations and Future Work.} 
While our identification of navigation-specific attention heads enables cost-free anomaly detection and recovery, merely correcting the physical trajectory without synchronizing the VLA model’s internal context can trigger repetitive cognitive errors. Implementing active replanning or re-prompting frameworks to align these internal states would effectively break such error loops, thereby significantly enhancing overall navigation success rates. 
Furthermore, within the context of prior research on pruning redundant attention heads for model compression~\cite{michel2019sixteen}, our identification of navigation heads offers a strategic pathway for targeted optimization. By selectively retaining only the task-essential heads, inference speeds could be significantly accelerated, thereby enabling more efficient and responsive operation in on-device deployments.

%% file: tex/supplementary.tex
\setcounter{page}{1}
\setcounter{section}{0}
\renewcommand{\thesection}{\Alph{section}}
\renewcommand{\theHsection}{\Alph{section}}
\providecommand{\theHpage}{supp.\arabic{page}}

\vspace{5\baselineskip}

\begin{center}
    {\Large \bfseries \boldmath Your Vision-Language-Action Model Already Has Attention Heads For Path Deviation Detection \par}
    
    \vspace{1.5em}
    
    {\large Supplementary Material \par}
\end{center}

\begin{figure*}[!h]
\vspace{20pt}
\centering
\includegraphics[width=\textwidth]{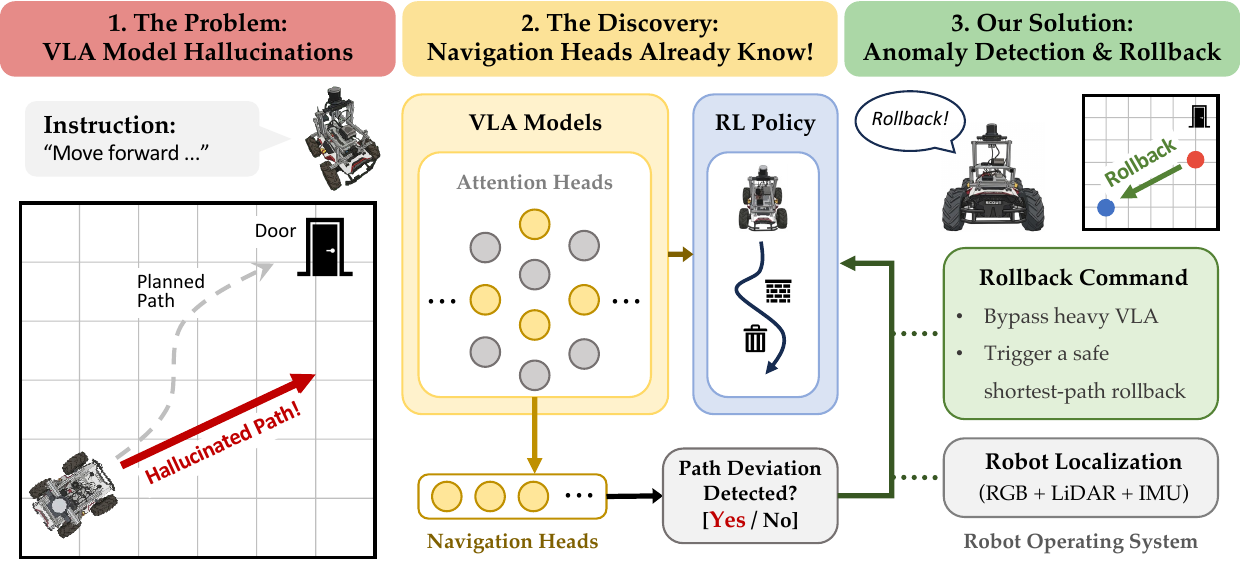}
\end{figure*}
\vspace{10pt}

\section*{Contents}

\noindent
\hyperref[sup:a]{\makebox[2em][l]{A} Expanding Path Deviation Detection \dotfill 2} \\[0.5ex]

\noindent
\hyperref[sup:b]{\makebox[2em][l]{B} Obstacle Avoidance RL Action Policy \dotfill 3} \\[0.5ex]
\hspace*{2em}\hyperref[sup:b1]{\makebox[2.5em][l]{B.1} IsaacLab Simulation Setup \dotfill 3} \\[0.5ex]
\hspace*{2em}\hyperref[sup:b2]{\makebox[2.5em][l]{B.2} Dynamic 2D Costmap Observation \dotfill 5} \\[0.5ex]
\hspace*{2em}\hyperref[sup:b3]{\makebox[2.5em][l]{B.3} Reinforcement Learning Configuration \dotfill 7} \\[0.5ex]
\hspace*{2em}\hyperref[sup:b4]{\makebox[2.5em][l]{B.4} RL Policy Computational Resources \dotfill 9} \\[0.5ex]
\hspace*{2em}\hyperref[sup:b5]{\makebox[2.5em][l]{B.5} Performance Metrics by Target Distance \dotfill 10} \\[0.5ex]
\hspace*{2em}\hyperref[sup:b6]{\makebox[2.5em][l]{B.6} Real-World Reactive Navigation of the RL Policy \dotfill 10} \\[1.5ex]

\noindent
\hyperref[sup:c]{\makebox[2em][l]{C} Real-Robot Platform System Implementation \dotfill 13} \\[0.5ex]
\hspace*{2em}\hyperref[sup:c1]{\makebox[2.5em][l]{C.1} System Specifications \dotfill 13} \\[0.5ex]
\hspace*{2em}\hyperref[sup:c2]{\makebox[2.5em][l]{C.2} Kinematic Action Smoother \dotfill 16} \\[0.5ex]
\hspace*{2em}\hyperref[sup:c3]{\makebox[2.5em][l]{C.3} State Estimation Comparison \dotfill 17} \\[1.5ex]

\noindent
\hyperref[sup:d]{\makebox[2em][l]{D} Real-World Deployment Results \dotfill 19} \\[0.5ex]

\clearpage
\section{Expanding Path Deviation Detection}
\label{sup:a}

\begin{table}[!ht]
\centering
\renewcommand{\arraystretch}{1.3}
\setlength{\tabcolsep}{5pt}
\caption{Computational overhead analysis on a single NVIDIA RTX 6000 Ada GPU. The integration of our path deviation detection module introduces a marginal latency increase of less than 20 ms per step, with no additional GPU memory required.}
\resizebox{0.75\columnwidth}{!}{
\begin{tabular}{l | ccc}
\hlineB{2.5}
\textbf{Configuration} & \textbf{Total Time} & \textbf{Alloc VRAM} & \textbf{Peak VRAM} \\
\hline
NaVILA (Baseline) & 583.5\,ms & 17,145.6 \,MB & 18,730.3 \,MB \\
NaVILA + Ours     & 603.3\,ms & 17,145.6 \,MB & 18,730.3 \,MB \\
\hline
\textbf{Difference} & \textbf{+19.8} \,ms & \textbf{0.0} \,MB & \textbf{0.0} \,MB \\
\hlineB{2.5}
\end{tabular}}
\vspace{10pt}
\label{sup:tab_1}
\end{table}

Expanding upon Sec. 4.2 of the main paper, Table~\ref{sup:tab_1} demonstrates the near-zero computational overhead of our path deviation detection mechanism. This exceptional efficiency stems from repurposing the attention weights, which are inherently computed as a natural byproduct of the standard LVLM forward pass. 
Provided that the task-relevant navigation heads ($H_\texttt{nav}$) and the optimal natural threshold ($\tau$) are pre-identified via offline statistical analysis, the online inference phase merely performs an $\mathcal{O}(1)$ lookup to extract specific attention distributions from $H_\texttt{nav}$. 
Because the subsequent detection logic—comprising entropy computation over visual tokens, relative difference calculation against a rolling-window mean, and threshold comparison—consists entirely of lightweight scalar operations, the mechanism requires no additional neural network computation beyond what the model already performs.
Furthermore, to complement the results in Table 2 of the main paper, we conducted additional evaluations on the RxR dataset~\cite{ku2020room} to assess the generalizability of our approach in environments with greater linguistic and spatial complexity. As shown in Table~\ref{sup:tab_2}, we followed the same offline analysis protocol, identifying the $H_\texttt{nav}$ exclusively from the RxR training set. These configurations were subsequently frozen and evaluated on the Val-Seen and Val-Unseen splits.
The results demonstrate that our method delivers robust performance across both splits. 
Notably, even in the RxR Val-Unseen environment, the proposed module yielded an episode-level EDR of 32.7\,\% and an FER of 9.7\,\%, while maintaining a remarkably high step-level precision of 93.2\,\%. These results confirm that our attention-based deviation detection serves as a robust, dataset-agnostic indicator of navigational confidence.

\begin{table}[!b]
\vspace{-15pt}
\caption{
    \textbf{Path Deviation Detection Performance on the RxR Dataset.} Extending the primary evaluation on the R2R dataset (Table 2 of the main paper), we further validate our proposed method on the RxR dataset~\cite{ku2020room}. 
    The detection mechanism, configured with $H_\texttt{nav}$ and hyperparameters derived exclusively from the training set, demonstrates robust generalizability across RxR validation splits, consistently maintaining high precision and significant EDR-FER gaps.
}
\vspace{-5pt}
\centering
\renewcommand{\arraystretch}{1.4}
\setlength\tabcolsep{2.4pt}
\small
\adjustbox{max width=\columnwidth}{
    \begin{tabular}{l|ccc|ccc|ccc|ccc}
        \hlineB{2.5}
        \multirow{3}{*}{\textbf{Method}} & \multicolumn{6}{c|}{\textbf{RxR Val-Seen}} & \multicolumn{6}{c}{\textbf{RxR Val-Unseen}} \\
        \cline{2-13}
         & \multicolumn{3}{c|}{\textbf{All Episodes}} & \multicolumn{3}{c|}{\textbf{$\mathcal{N} \to \mathcal{A}$ Step-level}} & \multicolumn{3}{c|}{\textbf{All Episodes}} & \multicolumn{3}{c}{\textbf{$\mathcal{N} \to \mathcal{A}$ Step-level}} \\
        \cline{2-13}
         & \textbf{EDR} $\uparrow$ & \textbf{FER} $\downarrow$ & \textbf{Gap} $\uparrow$ & \textbf{Prec.} $\uparrow$ & \textbf{Rec.} $\uparrow$ & \textbf{F1} $\uparrow$ & \textbf{EDR} $\uparrow$ & \textbf{FER} $\downarrow$ & \textbf{Gap} $\uparrow$ & \textbf{Prec.} $\uparrow$ & \textbf{Rec.} $\uparrow$ & \textbf{F1} $\uparrow$ \\
        \hline
        \textbf{Ours} & 36.1\,\% & 6.7\,\% & 29.4\,\% & 96.9\,\% & 39.8\,\% & 56.5\,\% & 32.7\,\% & 9.7\,\% & 23.0\,\% & 93.2\,\% & 34.4\,\% & 50.3\,\% \\
        \hlineB{2.5}
    \end{tabular}
}
\vspace{-5pt}
\label{sup:tab_2}
\end{table}

\clearpage
\section{Obstacle Avoidance RL Action Policy}
\label{sup:b}

\subsection{IsaacLab Simulation Setup}
\label{sup:b1}

\begin{figure*}[!h]
\centering
\includegraphics[width=0.93\textwidth]{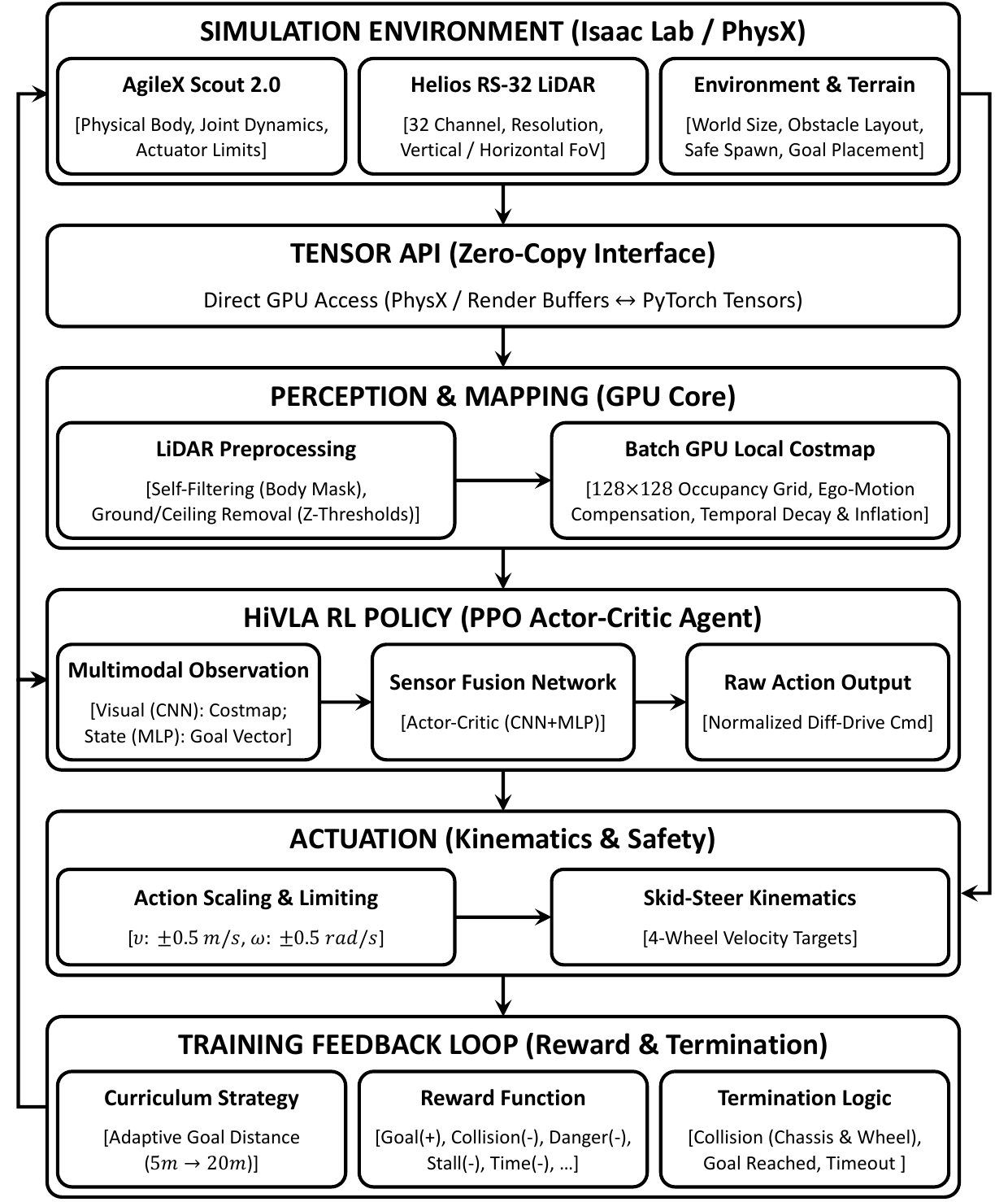}
\caption{
\textbf{RL training diagram.}
Utilizing IsaacLab~\cite{mittal2025isaaclab} and PhysX, we construct a simulation environment that precisely aligns with real-world dynamics, incorporating the URDF-defined robot geometry and sensor extrinsics while tuning skid-steer friction, damping, and actuator forces alongside the 32-channel LiDAR configurations. 
To facilitate reward discovery and ensure stable convergence, we employ a curriculum strategy that adaptively scales the navigation challenge by increasing the goal distance from 5\,m to 20\,m across 20 sub-terrains mixed with cylinders and walls, as illustrated in Fig. 9 of the main paper. 
With rewards tailored for collision avoidance and goal-reaching, this large-scale training of 1,024 agents on a single NVIDIA RTX 6000 Ada GPU enables comprehensive exploration of the state space and robust policy optimization.
}
\label{sup:fig_1}
\end{figure*}

\clearpage

\begin{table}[!t]
\centering
\renewcommand{\arraystretch}{1.5}
\setlength{\tabcolsep}{2pt}
\caption{\textbf{Configuration of the IsaacLab~\cite{mittal2025isaaclab} Simulation Environment}}
\resizebox{\columnwidth}{!}{%
\begin{tabular}{llc}
\hlineB{2.5}
\textbf{Category} & \textbf{Parameter} & \textbf{Value} \\ 
\hline
\multirow{4}{*}{Robot \& Kinematics} 
 & Wheel Radius ($R$) / Wheelbase ($L$) & 0.165 m / 0.582 m \\
 & Robot Dimensions (Length $\times$ Width) & 1.06 m $\times$ 0.90 m \\
 & Max Linear / Angular Velocity & ±0.5 m/s / ±0.5 rad/s \\
 & Policy Control Frequency & 15 Hz ($\approx$0.067 s/step) \\ 
\hline
\multirow{4}{*}{3D LiDAR Sensor} 
 & Sensor Mounting Height & 0.63 m \\
 & Channels / Max Range & 32 / 10.0 m \\
 & Vertical / Horizontal FOV & -55° to +15° / 360° \\
 & Horizontal Resolution & 0.2° (57,632 rays/scan) \\
\hline
\multirow{5}{*}{Terrain \& Environment} 
 & Sub-terrains (Size / Count) & 60 m $\times$ 60 m / 20 tiles \\
 & Cylinder Obstacles & 80 per tile (Radius: 0.25 m, Height: 2.0 m) \\
 & Wall Obstacles & 60 per tile (1.5 m $\times$ 0.75 m $\times$ 2.0 m) \\
 & Min Obstacle Clearance & 2.0 m \\
 & Min Spawn Clearance & 1.5 m \\ 
\hline
\multirow{4}{*}{Termination Conditions} 
 & Goal Reached Threshold & 0.5 m \\
 & Timeout Limit & 150.0 s \\
 & Collision Threshold (Chassis / Wheels) & $>$ 200.0 N \\
 & Stall Condition & $<$ 0.2 m displacement in 3.0 s \\ 
\hlineB{2.5}
\end{tabular}%
}
\vspace{-10pt}
\label{sup:tab_3}
\end{table}

To facilitate robust sim-to-real transfer, we meticulously designed the RL training environment within IsaacLab~\cite{mittal2025isaaclab}, as outlined in Fig.~\ref{sup:fig_1}. Crucially, this simulation is engineered to precisely mirror our real-world physical setup, as demonstrated in Fig.~\ref{sup:fig_4}. The comprehensive list of physical and environmental parameters utilized for this configuration is detailed in Table~\ref{sup:tab_3}.

\myparagraph{Robot Configuration and Kinematics.} To minimize the reality gap, we constructed a custom USD model of our mobile robot and imported it into the simulation at a 1:1 scale. Because the physical robot operates on a skid-steer differential drive mechanism, we explicitly engineered the left and right wheel velocities to be governed by the kinematic equations $v_{left} = (v - \omega \cdot L / 2) / R$ and $v_{right} = (v + \omega \cdot L / 2) / R$, where $v$ and $\omega$ denote the linear and angular velocities, respectively. We strictly applied the real-world specifications: a wheelbase of $L = 0.582 \text{ m}$ and a wheel radius of $R = 0.165 \text{ m}$. The RL policy utilizes a $\tanh$ activation to output continuous actions bounded within $[-1, 1]$, which we explicitly mapped to a maximum linear velocity of $0.5 \text{ m/s}$ and a maximum angular velocity of $0.5 \text{ rad/s}$. To define the physical collision boundaries during navigation, we adopted a conservative rectangular robot dimension of $1.06 \text{ m} \times 0.90 \text{ m}$, incorporating an empirical safety margin over the bare chassis measured directly on the physical robot. To balance rapid reactive obstacle avoidance with computational efficiency while maintaining physical realism, we designed the policy to operate at a control frequency of $15 \text{ Hz}$ (\ie, $\approx 0.067 \text{ s}$ per step), allowing a maximum translational displacement of $\approx 0.033 \text{ m}$ per action step, while the underlying physics engine simulates dynamics at $60 \text{ Hz}$.

\myparagraph{Sensor Setup.} To guarantee seamless sim-to-real transfer of perception, we configured the 3D LiDAR module provided by IsaacLab to faithfully replicate the mounting position and hardware specifications of the physical Helios RS-32 LiDAR used in our real-world deployments. Specifically, the simulated sensor is mounted at a height of $0.63 \text{ m}$ above the ground. It uses 32 vertical channels to enforce a vertical field of view (FOV) from $-55^\circ$ to $+15^\circ$, along with a full $360^\circ$ horizontal FOV. With a $0.2^\circ$ horizontal resolution, the sensor casts 57,632 rays per scan up to a maximum range of $10.0 \text{ m}$, providing a dense geometric representation of the surroundings identical to the real robot's sensory input.

\myparagraph{Environment and MDP Definitions.} Training occurs across a massive procedurally generated terrain composed of 20 sub-terrains. To enforce a reproducible and physically viable evaluation standard, we strictly populated each sub-terrain tile with 80 cylinders (radius $0.25 \text{ m}$, height $2.0 \text{ m}$) and 60 wall-shaped boxes ($1.5 \text{ m} \times 0.75 \text{ m} \times 2.0 \text{ m}$). We configured a rejection-sampling algorithm to position the obstacles, ensuring a minimum obstacle-to-obstacle clearance of $2.0 \text{ m}$, with a $3.0 \text{ m}$- wide obstacle-free platform deliberately preserved at the center of each tile. At episode initialization, we programmed the robot to spawn $0.5 \text{ m}$ above the ground, with a randomized yaw orientation to prevent physics clipping, and to enforce a strict $1.5 \text{ m}$ obstacle-free spawn clearance. We defined the Markov Decision Process (MDP) to terminate under four distinct conditions: (i) reaching within $0.5 \text{ m}$ of the goal, (ii) exceeding the maximum time limit of $150.0 \text{ s}$, (iii) experiencing physical collisions exceeding $200.0 \text{ N}$ on either the chassis or the wheels (after a brief 20-step warm-up period to ignore initial spawn impacts), or (iv) stalling (trapped within a $0.2 \text{ m}$ radius for over $3.0 \text{ s}$).

\subsection{Dynamic 2D Costmap Observation}
\label{sup:b2}

\begin{figure*}[!t]
\centering
\includegraphics[width=0.95\textwidth]{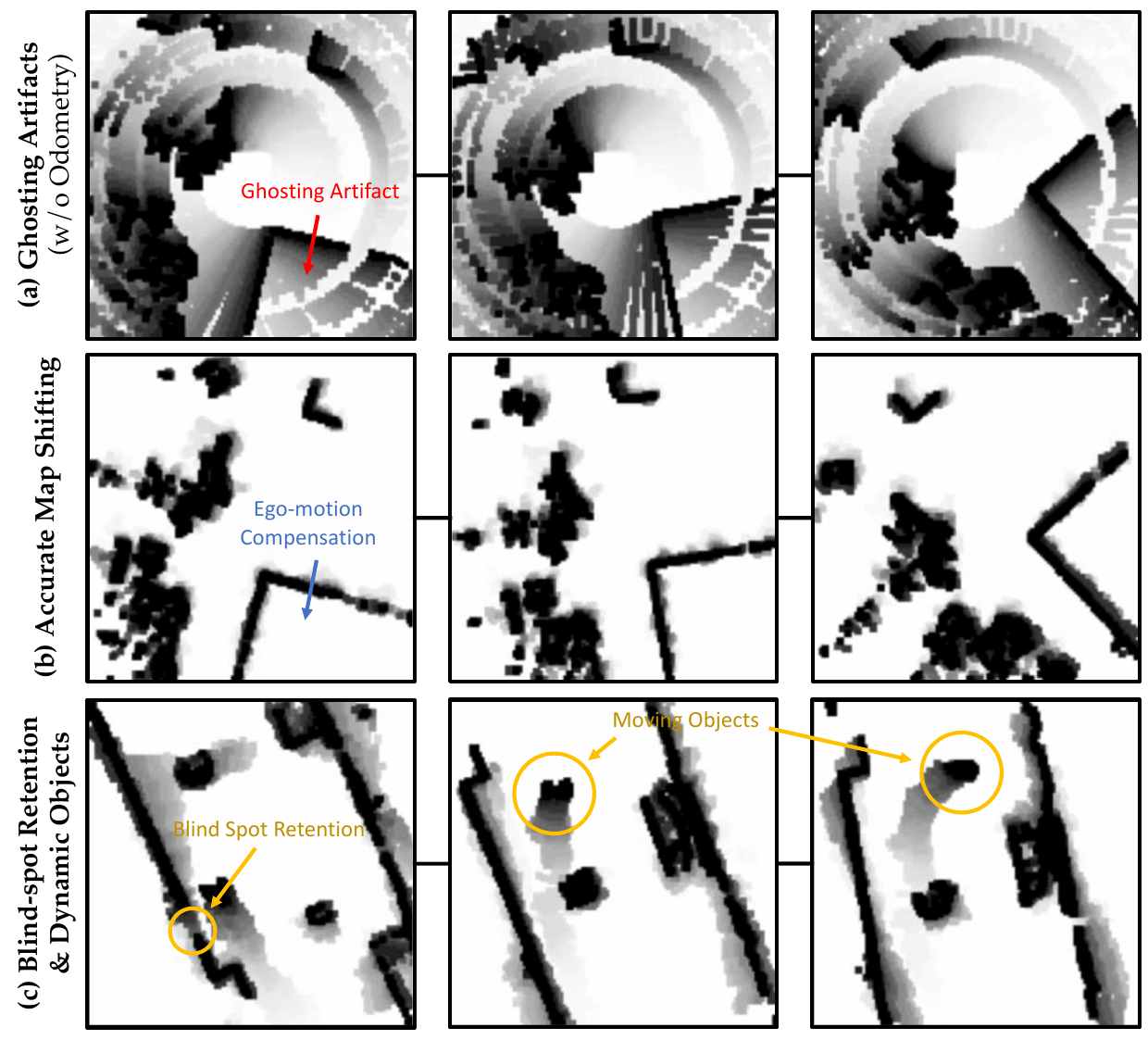}
\caption{
    \textbf{Dynamic 2D Costmap Observation with Ego-Motion Compensation and Temporal Decay.} (a) Without proper odometry, uncompensated ego-motion causes severe ghosting artifacts during a clockwise rotation, permanently corrupting the spatial memory. (b) By incorporating strict ego-motion compensation during the same rotation, the local costmap accurately shifts in accordance with the robot's movement, maintaining a clean and consistent representation. (c) Applying an optimal temporal decay factor ($\gamma = 0.9$) allows the system to temporarily retain static obstacles even when they enter the sensor's blind spot, while gracefully fading the historical trails of dynamic moving objects.
}
\vspace{-20pt}
\label{sup:fig_2}
\end{figure*}

Rather than relying exclusively on sparse 1D LiDAR arrays, our RL policy operates on a dynamic 2D occupancy grid costmap. This intermediate representation, as illustrated in Fig.~\ref{sup:fig_2}, equips the agent with spatial memory without requiring Recurrent Neural Networks (RNNs) and ensures sim-to-real robustness through sensor agnosticism. While raw 1D data is easily overfitted to specific hardware specifications, the 2D costmap decouples the input space from the raw sensor stream, allowing the policy to remain fully compatible with different LiDAR models or depth cameras without retraining. The visual observation space is a single-channel $128 \times 128$ tensor that captures a $6.4 \text{ m} \times 6.4 \text{ m}$ physical area centered on the robot, yielding a dense, uniform $0.05 \text{ m/pixel}$ resolution.

\myparagraph{Hardware-Accelerated Pipeline and Sim-to-Real Alignment.} Processing 57,632 LiDAR rays on the CPU for thousands of parallel agents in simulation induces a severe computational bottleneck. To resolve this, we constructed the entire mapping pipeline using CUDA-optimized PyTorch operations and executed it directly on the GPU in FP16 precision. To eliminate the sim-to-real gap in the perception module, this identical PyTorch codebase is deployed on the physical robot's NVIDIA Jetson AGX Orin. By setting the batch size to 1, the physical robot leverages its native Tensor Cores to process the costmap in real time with the same logic, completely bypassing CPU-GPU transfer overheads.

\myparagraph{Ego-Motion Compensation via Zero-Allocation Warping.} To maintain a globally consistent map, the local costmap must accurately shift in response to the robot's movement. Our pipeline integrates 3D LiDAR point clouds with high-frequency odometry data to compute the relative translational and rotational displacements at each policy step. To eliminate overhead during real-time execution, we pre-allocate an affine transformation matrix buffer, fundamentally eradicating memory allocation and Garbage Collection (GC) spikes during navigation. Subsequently, we warp the previous costmap using sub-pixel grid sampling (\texttt{F.grid\_sample}) to apply the inverse of the robot's motion. During this process, newly revealed unknown areas entering the field of view are strictly zero-padded to ensure safety.

\myparagraph{Temporal Decay for Dynamic Environments.} Real-world environments inherently contain moving objects. If the costmap remains static (i.e., a decay factor of 1.0), a passing pedestrian would leave a permanent obstacle trail, completely blocking the robot's path. Conversely, decaying the map too rapidly causes the agent to instantly forget static obstacles in temporary blind spots, leading to collisions. To resolve this dilemma, we apply a temporal decay factor of $\gamma = 0.9$ at each step. This specific rate strikes an optimal balance for spatial memory—safely remembering temporary blind spots while gracefully fading the historical trails of dynamic entities.

\myparagraph{Spatial Filtering and Safety Margins.} Finally, to ensure the policy reacts exclusively to genuine navigational hazards, we filter the LiDAR points along the Z-axis, retaining only data within $0.1 \text{ m} \le Z \le 1.0 \text{ m}$ to mask out ground noise and overhead structures. Furthermore, we apply a geometric self-masking filter to prevent the robot from perceiving its own chassis or antennas as obstacles. Rather than relying on simple CAD dimensions, the boundaries of this $1.10 \text{ m} \times 0.80 \text{ m}$ exclusion footprint were empirically derived by directly measuring the physical RS-LiDAR sensor's self-occlusion view. Any points falling within this measured zone are strictly ignored. To prevent the policy from shaving corners too closely, we apply a 2D Max Pooling operation ($3 \times 3$ kernel) over the grid, geometrically inflating all valid obstacles by an exact 1-pixel ($0.05 \text{ m}$) safety margin with negligible computational overhead.

\subsection{Reinforcement Learning Configuration}
\label{sup:b3}

We utilize Proximal Policy Optimization (PPO) to train the navigation policy within the IsaacLab environment. The training configuration is structurally divided into the design of a dense reward function and the specification of PPO network hyperparameters, as detailed in Tables~\ref {sup:tab_4} and~\ref {sup:tab_5}, respectively.

\myparagraph{Reward Function.} To cultivate robust, human-like navigation in highly constrained environments, we engineered a dense reward function ($r_{\text{total}}$) structured into four domains (Table~\ref{sup:tab_4}). To formalize the robot's spatial perception, we define $d_{\text{corridor}}$ as the minimum obstacle distance within a forward safety corridor, flagging the path as blocked ($\mathbb{I}_{\text{blk}} = 1$) if $d_{\text{corridor}} < 0.8\text{ m}$. We calculate the lateral tightness ($\tau_{\text{side}} = \max(\tau_L, \tau_R) \in [0, 1]$) to quantify the narrowness of the immediate surroundings. The rationale for each component is as follows:
(i) \textbf{Task Progression:} Relying solely on a sparse success bonus ($r_{\text{success}}$) granted at a $0.5\text{ m}$ threshold is inefficient for long-horizon navigation. We employ a potential-based shaping reward ($r_{\text{goal}}$) utilizing an exponential function of the 2D goal distance ($d_t$). This provides a steepening, continuous gradient, effectively preventing the agent from being trapped in local minima. A step-wise time penalty ($r_{\text{time}}$) further encourages efficient routing.
(ii) \textbf{Kinematic Control:} To ensure dynamically feasible trajectories, the projected velocity reward ($r_{\text{vel}}$) incentivizes forward motion ($v_x$) aligned with the goal ($\cos(\psi_{\text{err}})$). Crucially, scaling this reward by $\tau_{\text{side}}$ and a distance-based speed factor ($s_{\text{speed}}$) encourages the policy to accelerate aggressively in open spaces while decelerating cautiously in narrow aisles. This is complemented by continuous heading alignment ($r_{\text{head}}$), step-wise yaw error reduction ($r_{\text{yaw}}$), and micro-displacement ($r_{\text{move}}$) rewards to maintain momentum without lateral drifting.
(iii) \textbf{Safety Enforcement:} For reliable sim-to-real transfer, a heavy discrete penalty ($r_{\text{col}}$) is strictly applied for physical collisions exceeding $200\text{ N}$ of force on the chassis or wheels. To proactively deter reckless driving, we introduce a proximity danger penalty ($r_{\text{danger}}$) formulated as a quadratic function of $d_{\text{corridor}}$. This establishes a virtual repulsive field that punishes dangerous approaches long before physical collisions occur.
(iv) \textbf{Behavioral Regularization:} Skid-steer robots often suffer from oscillatory freezing in tight spaces. We penalize excessive angular velocity ($r_{\text{ang}}$) in open areas and punish prolonged stalls ($r_{\text{stall}}$) exceeding a $1.0\text{ s}$ grace period. Most importantly, a centering reward ($r_{\text{center}}$) leverages the spatial imbalance between the left and right clearances ($\tau_R - \tau_L$) to naturally steer the robot toward the wider side, keeping it smoothly centered through bottlenecks.

\begin{table}[!t]
\centering
\renewcommand{\arraystretch}{1.3}
\setlength{\tabcolsep}{5pt}
\caption{\textbf{Dense Reward Function Configuration.} The total reward is the sum of these 12 components. $\mathbb{I}_{\text{blk}}$ denotes the path-blocked indicator ($d_{\text{corridor}} < 0.8\text{ m}$), and we define a speed scaling factor $s_{\text{speed}} = \text{clamp}(d_{\text{corridor}}/2.0, 0, 1)$.}
\resizebox{\columnwidth}{!}{%
\begin{tabular}{l l l}
\hlineB{2.5}
\textbf{Category} & \textbf{Component} & \textbf{Formulation \& Condition} \\
\hline
\multirow{3}{*}{\textbf{Task}} 
 & Goal Progress ($r_{\text{goal}}$) & $100 \cdot (e^{-0.25 d_t} - e^{-0.25 d_{t-1}})$ \\
 & Goal Success ($r_{\text{success}}$) & $+20.0$ \quad if $d_t <$ 0.5 m \\
 & Time Penalty ($r_{\text{time}}$) & $-0.05$ \quad per step \\
\hline
\multirow{4}{*}{\textbf{Kinematics}} 
 & Projected Velocity ($r_{\text{vel}}$) & $3.0 v_x \cos(\psi_{\text{err}}) (1 - 0.85 \tau_{\text{side}}) s_{\text{speed}} (1 - \mathbb{I}_{\text{blk}})$ \\
 & Heading Alignment ($r_{\text{head}}$) & $1.0 \cdot (1 - |\psi_{\text{err}}|/\pi) \cdot \mathbb{I}(|v_x| > 0.1) \cdot (1 - \mathbb{I}_{\text{blk}})$ \\
 & Yaw Error Reduction ($r_{\text{yaw}}$) & $2.0 \cdot (|\psi_{\text{err}}|_{t-1} - |\psi_{\text{err}}|_t)$ \\
 & Micro-Displacement ($r_{\text{move}}$) & $0.2 \cdot \tanh(\Delta s / 0.03)$ \\
\hline
\multirow{2}{*}{\textbf{Safety}} 
 & Collision Penalty ($r_{\text{col}}$) & $-30.0$ \quad if force $>$ 200 N (chassis or wheels) \\
 & Proximity Danger ($r_{\text{danger}}$) & $-3.0 \cdot (1 - \text{clamp}(d_{\text{corridor}} / 3.13, 0, 1))^2$ \\
\hline
\multirow{3}{*}{\textbf{Behavior}} 
 & Angular Penalty ($r_{\text{ang}}$) & $-0.3 \cdot |\omega_z| \cdot (1 - \tau_{\text{side}}) \cdot (1 - \mathbb{I}_{\text{blk}})$ \\
 & Centering Reward ($r_{\text{center}}$) & $2.5 \cdot (\tau_R - \tau_L) \cdot \omega_z$ \\
 & Stall Penalty ($r_{\text{stall}}$) & $\max(-0.5 \cdot \max(t_{\text{stall}} - 1.0, 0)^2, -2.0)$ \\
\hlineB{2.5}
\end{tabular}%
}
\label{sup:tab_4}
\end{table}

\myparagraph{PPO Hyperparameters.} Expanding upon the reinforcement learning methodology introduced in Sec. 3.5 of the main paper, Table~\ref{sup:tab_5} details the SKRL-based PPO configuration \cite{serrano2023skrl} for stabilizing 1,024 parallel IsaacLab environments. The specific rationale for each key hyperparameter is as follows:
(i) \textbf{Temporal Discounting and Advantage Estimation:} We deliberately lowered the discount factor to $\gamma = 0.96$ rather than the standard 0.99. In densely cluttered environments, distant future rewards are inherently uncertain. A lower $\gamma$ effectively shortens the agent's time horizon, thereby prioritizing immediate reactive obstacle avoidance over long-term planning. This temporal weighting is paired with a Generalized Advantage Estimation (GAE) parameter of $\lambda = 0.95$ to strike an optimal balance between bias and variance.
(ii) \textbf{Observation and Batch Processing:} We completely disabled the default state preprocessor. Since our primary visual observation is a structured 2D occupancy costmap, standard continuous normalization would distort its absolute geometric scale and distinct boundary values, thereby degrading the spatial memory of the CNN encoder. For policy optimization, collecting 96 rollout steps across all 1,024 environments yields a massive batch of 98,304 transitions per iteration. To maximize sample efficiency, these transitions are rigorously optimized over 5 epochs and 4 mini-batches.
(iii) \textbf{Action Space and Actor Architecture:} Driven by the non-holonomic kinematics of our target wheeled robot, the action space is explicitly restricted to two continuous control commands: the forward linear velocity ($v_x$) and the angular velocity ($\omega_z$). To ensure policy stability and prevent the generation of physically unbounded commands, the final output layer of the actor network employs a $\tanh$ activation function. This strictly bounds the raw model outputs to the normalized range $[-1, 1]$ and subsequently scales them to match the robot chassis's maximum physical velocity limits during simulation.
(iv) \textbf{Adaptive Learning Rate:} To maintain policy stability across diverse procedurally generated sub-terrains, we employ a KL-adaptive learning rate scheduler starting at a base rate of $3 \times 10^{-4}$. By dynamically scaling the learning rate based on a target KL divergence threshold of 0.008, we strictly bound the magnitude of policy updates. This mechanism prevents excessively large policy steps, which can lead to catastrophic forgetting, ensuring that the robot retains its critical collision-avoidance reflexes throughout 100,000 training timesteps.

\begin{table}[!t]
\centering
\renewcommand{\arraystretch}{1.3}
\setlength{\tabcolsep}{5pt}
\caption{\textbf{PPO Network and Training Hyperparameters} (SKRL~\cite{serrano2023skrl})}
\resizebox{0.9\columnwidth}{!}{%
\begin{tabular}{lc|lc}
\hlineB{2.5}
\textbf{Hyperparameter} & \textbf{Value} & \textbf{Hyperparameter} & \textbf{Value} \\
\hline
Algorithm & PPO & Learning Rate & $3 \times 10^{-4}$ \\
Parallel Environments & 1,024 & LR Scheduler & KL-Adaptive (0.008) \\
Rollout Length & 96 steps & Discount Factor ($\gamma$) & 0.96 \\
Learning Epochs & 5 & GAE Parameter ($\lambda$) & 0.95 \\
Mini-batches & 4 & Total Timesteps & 100,000 \\
State Preprocessor & Disabled & Checkpoint Interval & 5,000 steps \\
\hlineB{2.5}
\end{tabular}%
}
\label{sup:tab_5}
\end{table}

\subsection{RL Policy Computational Resources}
\label{sup:b4}

\begin{table}[!ht]
\centering
\renewcommand{\arraystretch}{1.3}
\setlength{\tabcolsep}{5pt}
\caption{
    \textbf{Computational overhead analysis of the RL Action Policy.} Evaluated on a single NVIDIA RTX 6000 Ada GPU, the lightweight CNN + MLP architecture (4.89\,M parameters) achieves a near-instantaneous inference latency of 0.2\,ms per step with minimal VRAM utilization.
}
\resizebox{0.75\columnwidth}{!}{
\begin{tabular}{l | ccc}
\hlineB{2.5}
\textbf{Configuration} & \textbf{Total Time} & \textbf{Alloc. VRAM} & \textbf{Peak VRAM} \\
\hline
\textbf{Ours (RL Policy)} & 0.2\,ms & 26.9 \,MB & 27.9 \,MB \\
\hlineB{2.5}
\end{tabular}}
\vspace{10pt}
\label{sup:tab_6}
\end{table}

The RL Action Policy is designed as a streamlined CNN $+$ MLP architecture with approximately 4.89\,M parameters, specifically optimized for high-frequency reactive control. As detailed in Table~\ref{sup:tab_6}, this lightweight configuration enables ultra-low-latency inference, requiring only 0.2\,ms per step on an NVIDIA RTX 6000 Ada GPU. 
With a minimal VRAM footprint of less than 28\,MB, this RL policy provides the essential agility for real-time obstacle avoidance and stability. By maintaining a negligible computational load even during continuous high-rate actuation, it ensures overall system efficiency and ultimately guarantees a stable control frequency of 10\,Hz during real-world deployment, as detailed in Sec.~\ref{sup:c}.

\subsection{Performance Metrics by Target Distance}
\label{sup:b5}

\begin{table}[!ht]
\centering
\renewcommand{\arraystretch}{1.3}
\setlength{\tabcolsep}{3pt}
\caption{
    Navigation performance by distance. SR: Success Rate, CR: Collision Rate, TR: Timeout Rate.
}
\resizebox{\columnwidth}{!}{
\begin{tabular}{l ccc ccc ccc ccc}
\hlineB{2.5}
\textbf{Distance} & \multicolumn{3}{c}{5m} & \multicolumn{3}{c}{10m} & \multicolumn{3}{c}{15m} & \multicolumn{3}{c}{20m} \\
\cmidrule(lr){2-4} \cmidrule(lr){5-7} \cmidrule(lr){8-10} \cmidrule(lr){11-13}
\textbf{Method} & SR $\uparrow$ & CR $\downarrow$ & TR $\downarrow$ & SR $\uparrow$ & CR $\downarrow$ & TR $\downarrow$ & SR $\uparrow$ & CR $\downarrow$ & TR $\downarrow$ & SR $\uparrow$ & CR $\downarrow$ & TR $\downarrow$ \\
\hline
APF~\cite{khatib1986real} & 60.5\,\% & \underline{5.9\,\%} & 33.6\,\% & 48.4\,\% & \underline{10.0\,\%} & 41.6\,\% & \underline{41.4\,\%} & \textbf{10.0\,\%} & 48.6\,\% & \underline{39.3\,\%} & \textbf{8.8\,\%} & 51.9\,\% \\
DWA~\cite{fox2002dynamic} & 17.6\,\% & 75.8\,\% & 6.6\,\% & 16.2\,\% & 82.8\,\% & \textbf{1.0\,\%} & 21.1\,\% & 78.1\,\% & \textbf{0.8\,\%} & 16.2\,\% & 81.8\,\% & \underline{2.0\,\%} \\
MPPI~\cite{williams2017information} & 3.7\,\% & 90.0\,\% & \underline{6.3\,\%} & 2.7\,\% & 91.6\,\% & 5.7\,\% & 1.6\,\% & 91.6\,\% & 6.8\,\% & 2.0\,\% & 91.4\,\% & 6.6\,\% \\
TEB~\cite{rosmann2012trajectory} & \underline{65.6\,\%} & 24.0\,\% & 10.4\,\% & \underline{51.0\,\%} & 40.6\,\% & 8.4\,\% & 41.0\,\% & 50.4\,\% & 8.6\,\% & 31.1\,\% & 59.4\,\% & 9.6\,\% \\
\hline
\textbf{Ours} & \textbf{92.2\,\%} & \textbf{5.1\,\%} & \textbf{2.7\,\%} & \textbf{89.8\,\%} & \textbf{8.6\,\%} & \underline{1.6\,\%} & \textbf{83.6\,\%} & \underline{14.3\,\%} & \underline{2.1\,\%} & \textbf{83.9\,\%} & \underline{14.5\,\%} & \textbf{1.6\,\%} \\
\hlineB{2.5}
\vspace{5pt}
\end{tabular}}
\label{sup:tab_7}
\end{table}

Expanding upon Sec. 4.3 of the main paper, Table~\ref{sup:tab_7} details the performance across target distances from 5\,m to 20\,m to analyze long-horizon robustness beyond the average metrics. As distance increases, baselines exhibit significant degradation: APF’s termination rate (TR) rises to 51.9\,\% at 20\,m due to cumulative entrapment in local minima, while TEB’s collision rate (CR) climbs to 59.4\,\% in complex layouts.  DWA and MPPI consistently fail across all ranges with CR exceeding 75\,\%. In contrast, our policy maintains a Success Rate (SR) of 83.9\,\% even at the maximum distance, more than doubling the performance of the strongest baseline (APF, 39.3\,\%). Notably, despite the fourfold increase in distance, our TR remains exceptionally stable below 3.0\,\%, proving that the policy effectively avoids local minima and maintains consistent safety in dense environments where classical reactive planners frequently stall or collide.

\subsection{Real-World Reactive Navigation of the RL Policy}
\label{sup:b6}

\begin{figure*}[!ht]
\centering
\includegraphics[width=0.9\textwidth]{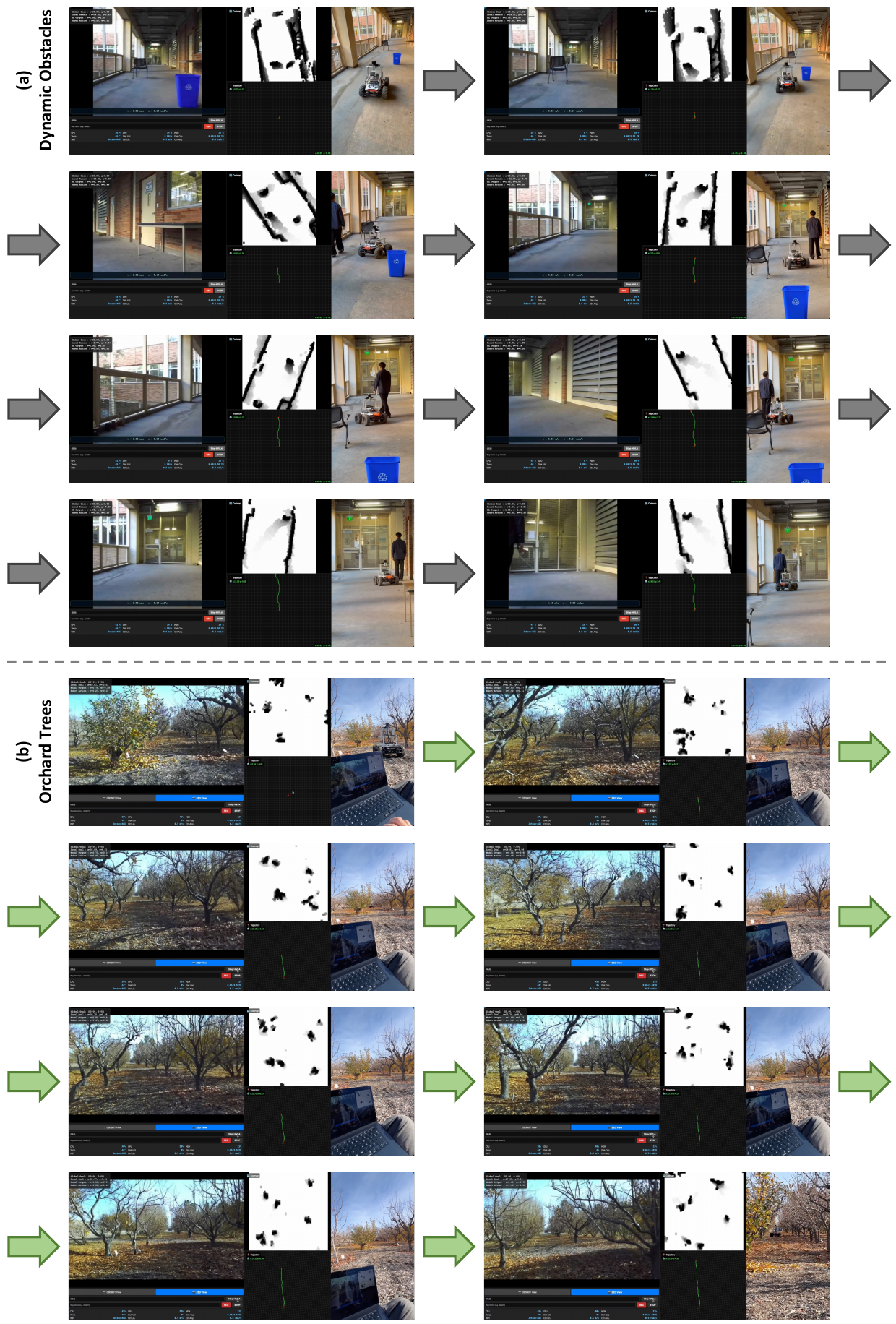}
\caption{
    \textbf{Real-world navigation performance of the standalone RL policy.} Operating at 10 Hz without VLA guidance, the agent safely navigates dynamic environments over tens of meters. \textbf{(a)} Indoors, it successfully avoids static objects and moving pedestrians. \textbf{(b)} In an outdoor orchard, it demonstrates robust collision avoidance against complex, irregular trees.
}
\label{sup:fig_3}
\end{figure*}

To evaluate the low-level action policy for obstacle avoidance, we demonstrated the real-world navigation performance of the standalone RL policy without VLA guidance. Given simple forward commands over distances of tens of meters, the agent operates at 10 Hz to safely navigate through dynamic and unstructured environments. As shown in Fig.~\ref{sup:fig_3}, (a) the policy successfully avoids both static objects and moving pedestrians indoors, and (b) it demonstrates robust collision avoidance against complex, irregularly shaped trees in an outdoor orchard farm.

\begin{figure*}[!ht]
\centering
\includegraphics[width=0.9\textwidth]{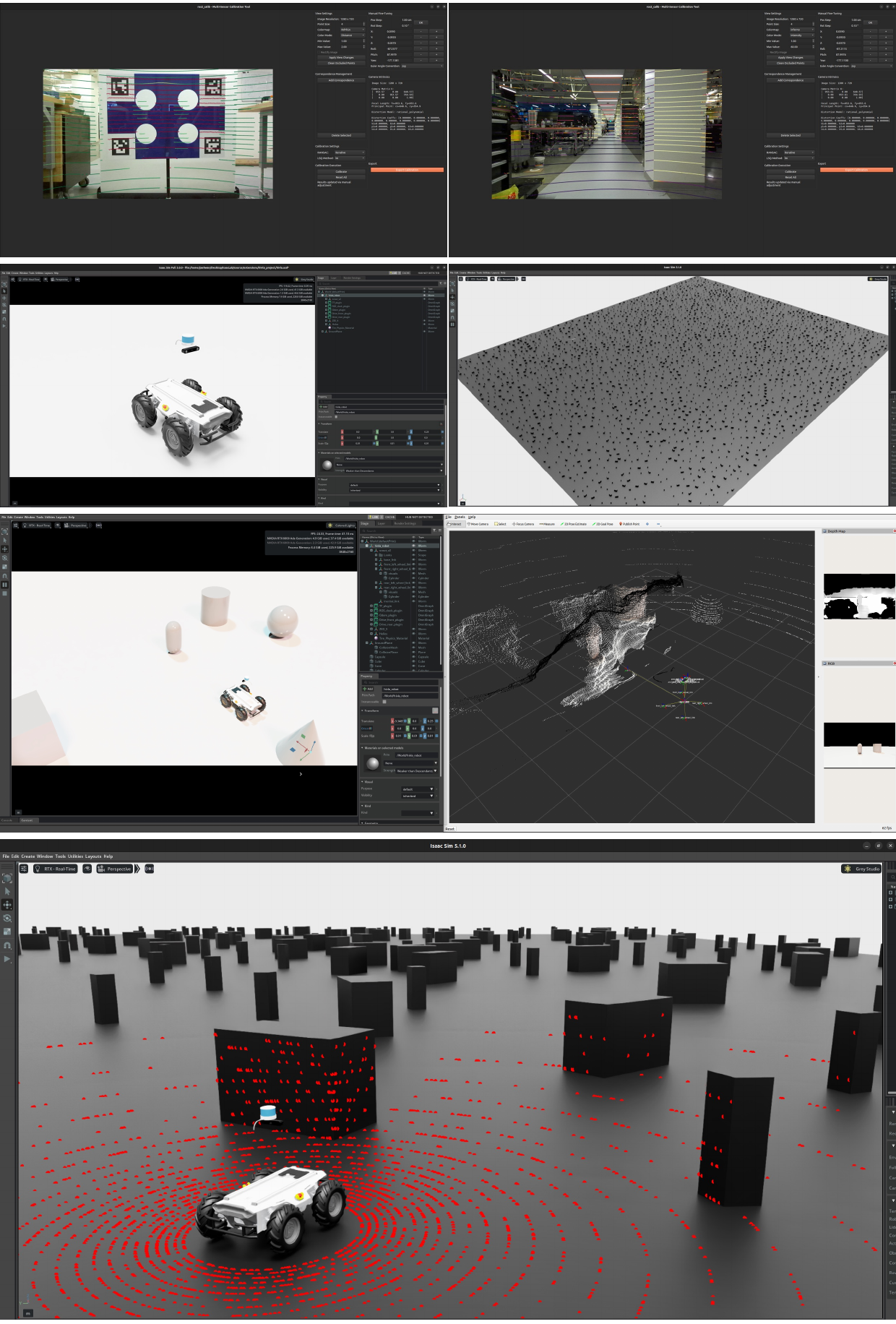}
\caption{
Overview of the robotic system and simulation environment.
\textbf{First row}: sensor calibration of the ZED 2i camera and RS-32 LiDAR. 
\textbf{Second row}: URDF robot model with sensor extrinsics and training terrains. 
\textbf{Third row}: Testing real-time ROS 2 integration within Isaac Sim.
\textbf{Fourth row}: training visualization of a single agent among 1,024 parallel instances.
}
\label{sup:fig_4}
\end{figure*}


\clearpage
\section{Real-Robot Platform System Implementation}
\label{sup:c}

\subsection{System Specifications}
\label{sup:c1}

\begin{figure*}[!h]
\centering
\includegraphics[width=\textwidth]{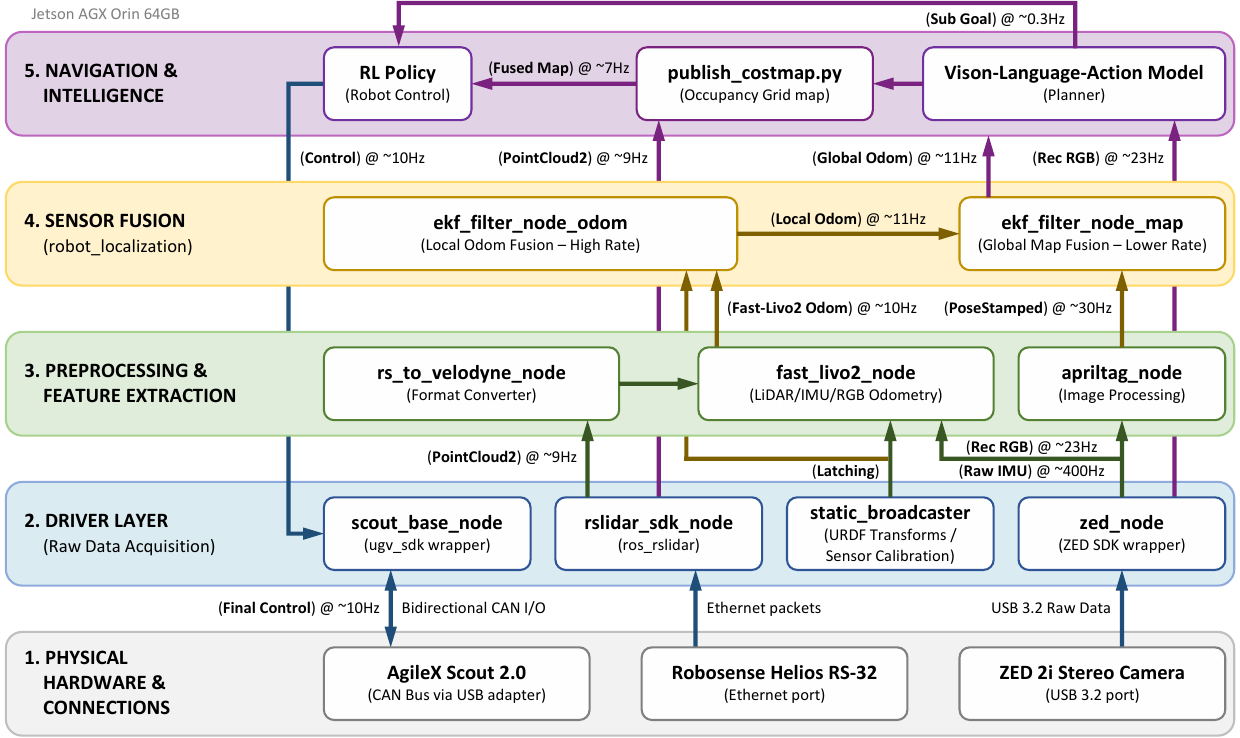}
\caption{ 
    \textbf{ROS 2 System Architecture.}
    The on-device pipeline integrates multimodal data from a skid-steer robot, a 32-channel spinning LiDAR, and a stereo RGB-D camera. For precise robot localization and efficient resource allocation, we employ Fast-LIVO2~\cite{zheng2024fast}, which leverages LiDAR, RGB, and IMU sensor streams. This is coupled with an Extended Kalman Filter (EKF) that fuses AprilTag~\cite{olson2011apriltag} detections to establish and maintain a robust globally fixed coordinate system. The navigation core operates asynchronously, comprising a $\sim$0.3\,Hz VLA planner for high-level sub-goal generation and a $\sim$10\,Hz RL policy that executes reactive control for obstacle avoidance.
}
\vspace{20pt}
\label{sup:fig_5}
\end{figure*}

\begin{figure*}[!t]
\centering
\includegraphics[width=\textwidth]{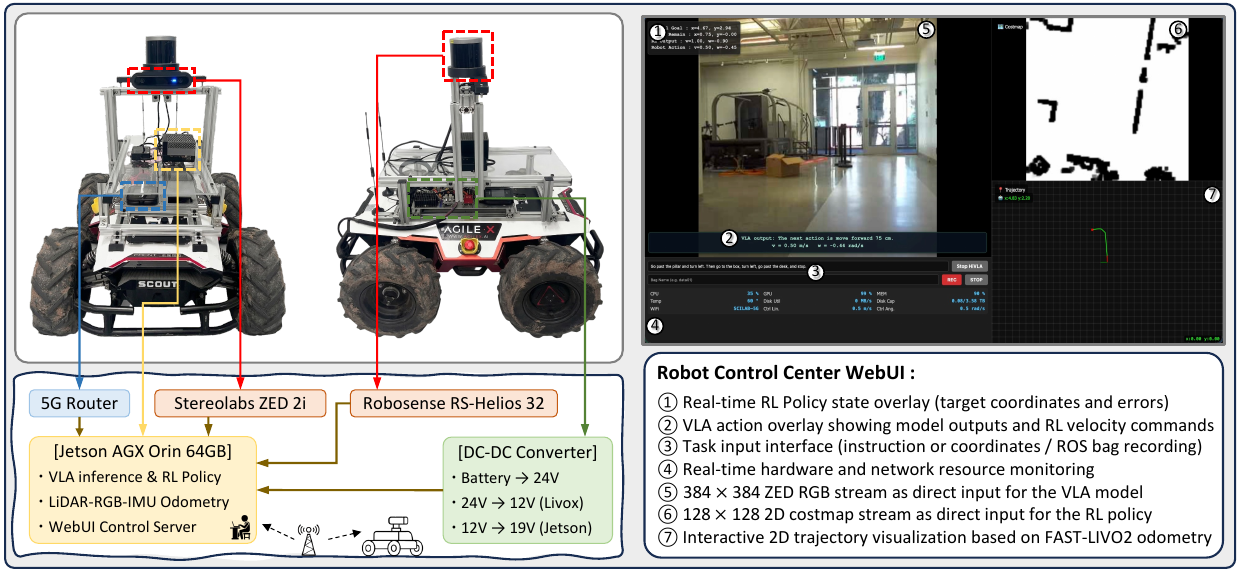}
\caption{
    \textbf{Physical Robot Setup and Control Center WebUI.} 
    (\textbf{Left}) Hardware configuration on the AgileX Scout 2.0 mobile base. A custom enclosure rigidly co-locates the Helios-RS 32 LiDAR and ZED 2i stereo camera. The onboard NVIDIA Jetson AGX Orin processes the VLA inference, RL policy, and Fast-LIVO2 odometry. Power distribution is managed via a dedicated DC-DC converter, while a 5G router ensures low-latency remote connectivity. 
    (\textbf{Right}) The custom Robot Control Center WebUI was deployed for teleoperation and monitoring.
}
\label{sup:fig_6}
\end{figure*}

Expanding upon Sec. 3.6 of the main paper, we developed a comprehensive hardware-software bridge on our custom skid-steer robot to successfully transfer the learned policies from simulation to the physical world. As shown in Fig.~\ref{sup:fig_5}, this architecture is explicitly designed to balance the high-frequency demands of reactive control with the intensive computational loads of visual reasoning. Furthermore, it enables effortless teleoperation and real-time monitoring of the robot through a dedicated WebUI, as illustrated in Fig.~\ref{sup:fig_6}.

\myparagraph{Software Setup.} To support this architecture, we configured a robust on-device environment, with detailed specifications summarized in Table~\ref{sup:tab_8}. The core compute engine is an NVIDIA Jetson AGX Orin 64GB running Ubuntu 22.04 and ROS 2 Humble. To maximize edge computing efficiency, the deep learning stack is fully optimized using JetPack 6.2.1 and CUDA 12.6. Within this ROS 2 ecosystem, the system effectively manages the asynchronous execution of multi-rate nodes. It seamlessly handles intensive inference for the 8-billion-parameter NaVILA model and attention-based path deviation detection, while simultaneously executing high-frequency Fast-LIVO2 sensor-fusion odometry and a lightweight RL action policy for reactive control on the AgileX Scout 2.0 mobile base. Furthermore, to facilitate real-time monitoring without degrading the main computational pipeline, we deployed a custom WebUI utilizing a Janus server and Flask. As shown in Fig.~\ref{sup:fig_6}, this integration ensures low-latency multimedia streaming and seamless teleoperation for the user.

\begin{table}[!t]
\centering
\renewcommand{\arraystretch}{1.3}
\setlength{\tabcolsep}{5pt}
\caption{
    \textbf{System specifications for physical deployment.} This table outlines the comprehensive hardware and software stack configured on the NVIDIA Jetson AGX Orin edge computer. It details the full hierarchical navigation pipeline, encompassing the 8-billion-parameter NaVILA model for high-level reasoning, the attention-based path deviation detection mechanism (parameterized by window size $W$ and thresholds), and the lightweight 4.89\,M parameter RL action policy for reactive local control.
}
\resizebox{0.9\columnwidth}{!}{
\begin{tabular}{lll}
\hlineB{2.5}
\textbf{Category} & \textbf{Component} & \textbf{Specification} \\
\hline
\multirow{4}{*}{Hardware} & Compute & NVIDIA Jetson AGX Orin Dev. 64\,GB \\
 & Mobile Base & AgileX Scout 2.0 \\
 & Camera & ZED 2i (Stereo RGB-D) \\
 & LiDAR & RoboSense Helios RS-32 \\
\hline
\multirow{6}{*}{System Software} & JetPack & 6.2.1 (L4T R36.4.7) \\
 & Kernel & 5.15.148-tegra \\
 & OS & Ubuntu 22.04 (aarch64) \\
 & CUDA & 12.6 \\
 & cuDNN & 9.3.0 \\
 & Python & 3.10.12 \\
\hline
ROS 2 Stack & ROS 2 Distro & Humble \\
\hline
\multirow{5}{*}{NaVILA Model} & Base Model & VILA (LLaMA-3 8B) \\
 & Vision Encoder & SigLIP-So400M/patch14/384 \\
 & Checkpoint & navila-llama3-8b-8f \\
 & Input Resolution & 384 $\times$ 384 \\
 & Inference Frames & 8 (7 history + 1 current) \\
 \hline
\multirow{4}{*}{Deviation Detection} 
 & Selected Heads ([L, H]) & [21, 12], [16, 1], [14, 1] \\
 & Window Size & $W = 10$ steps \\
 & Patience Threshold & $P = 9$ steps \\
 & Natural Threshold & $\tau = 0.95$ \\
 \hline
\multirow{4}{*}{RL Action Policy} & Architecture & CNN + MLP \\
 & Parameters & 4.89\,M \\
 & Costmap Input & $1 \times 128 \times 128$ (Occupancy Grid) \\
 & Goal Input & $1 \times 2$ ($\Delta x$, $\Delta y$ in local frame) \\
\hlineB{2.5}
\end{tabular}}
\label{sup:tab_8}
\end{table}

\clearpage
\myparagraph{Hardware Setup.} As visualized in Fig.~\ref{sup:fig_6}, the 32-channel Helios-RS 32 LiDAR is mounted centrally at a height of 0.63,m to maximize ground terrain visibility, while the ZED 2i stereo camera is positioned to provide an unobstructed egocentric view. This hardware configuration precisely aligns with the environment described in Sec.~\ref{sup:c}. To rigidly secure these sensors, we fabricated a custom hardware enclosure. Furthermore, to ensure accurate spatial alignment between the sensors, we performed cross-sensor extrinsic calibration using an interactive, targetless calibration tool~\cite{ros2_calib, jeong2025vision}. As depicted in Fig.~\ref{sup:fig_4}, this calibration process yields the following transformation matrix, $T_{\texttt{lidar}}^{\texttt{cam}}$:
\begin{align}
    T_{\texttt{lidar}}^{\texttt{cam}} &= 
    \begin{bmatrix}
         0.0017 & -0.0349 &  0.9994 &  0.0537 \\
        -1.0000 &  0.0017 &  0.0018 &  0.0593 \\
        -0.0017 & -0.9994 & -0.0349 & -0.0938 \\
         0      &  0      &  0      &  1 
    \end{bmatrix} \label{eq:extrinsic_matrix}
\end{align}
This precise spatial mapping plays a crucial role in deriving highly accurate results across the entire navigation pipeline: RGB-based path planning for the VLA model, LiDAR-based reactive control for the RL policy, and robust LiDAR-RGB-IMU fusion for the Fast-LIVO2 odometry.

\clearpage

\subsection{Kinematic Action Smoother}
\label{sup:c2}

\begin{table}[!h]
\centering
\renewcommand{\arraystretch}{1.3}
\setlength{\tabcolsep}{5pt}
\caption{\textbf{Kinematic Action Smoother Configuration.} Hyperparameters defining the physical constraints and symmetric acceleration limits for the skid-steer robot.}
\resizebox{0.6\columnwidth}{!}{%
\begin{tabular}{l c c}
\hlineB{2.5}
\textbf{Parameter} & \textbf{Symbol} & \textbf{Value} \\
\hline
Max Linear Velocity & $v_\texttt{max}$ & 0.5 m/s \\
Max Angular Velocity & $\omega_\texttt{max}$ & 0.5 rad/s \\
Linear Acceleration Limit & $a_\texttt{max}$ & 1.0 $\text{m/s}^2$ \\
Angular Acceleration Limit & $a_\texttt{max}^\omega$ & 5.0 $\text{rad/s}^2$ \\
Deadzone Threshold & $\epsilon$ & 0.02 \\
\hlineB{2.5}
\end{tabular}%
}
\vspace{15pt}
\label{sup:tab_9}
\end{table}

At each control step $t$ operating at 10 Hz ($\Delta t = 0.1\text{ s}$), the RL policy outputs target velocity commands $(v_t^\texttt{tgt}, \omega_t^\texttt{tgt})$. Directly applying these discontinuous commands to the physical robot can induce severe mechanical oscillations and physically unsafe accelerations. To ensure stable and safe locomotion, we implemented a three-stage Kinematic Action Smoother pipeline, utilizing the physical constraints and hyperparameters detailed in Table \ref{sup:tab_9}.

\noindent (i) \textbf{Velocity Saturation:} The raw target velocities are first strictly bounded to the physical limits of the robot chassis to prevent hardware overload:
\begin{align}
    \hat{v}_t = \text{clip}(v_t^\texttt{tgt}, -v_\texttt{max}, v_\texttt{max}), \quad \hat{\omega}_t = \text{clip}(\omega_t^\texttt{tgt}, -\omega_\texttt{max}, \omega_\texttt{max})
\end{align}

\noindent (ii) \textbf{Symmetric Acceleration Clamping:} We compute the instantaneous acceleration over the elapsed time $\Delta t$. To prevent jerky starts and stops, we enforce symmetric linear ($a_\texttt{max}$) and angular ($a_\texttt{max}^\omega$) acceleration limits. At the 10 Hz control frequency, this configuration allows the robot to safely decelerate from the maximum linear velocity ($v_\texttt{max}$) to a complete stop in exactly 5 steps (0.5 s) and execute a full angular direction reversal in 2 steps (0.2 s):
\begin{align}
    a_t &= \frac{\hat{v}_t - v_{t-1}}{\Delta t}, \quad &\bar{v}_t &= v_{t-1} + \text{clip}(a_t, -a_\texttt{max}, a_\texttt{max}) \cdot \Delta t \label{eq:acc_clamp_v} \\
    a_t^\omega &= \frac{\hat{\omega}_t - \omega_{t-1}}{\Delta t}, \quad &\bar{\omega}_t &= \omega_{t-1} + \text{clip}(a_t^\omega, -a_\texttt{max}^\omega, a_\texttt{max}^\omega) \cdot \Delta t 
\end{align}

\noindent (iii) \textbf{Deadzone Suppression:} Finally, sub-threshold commands are explicitly zeroed to suppress motor chatter and prevent mechanical creep when the robot is expected to remain stationary:
\begin{align}
    v_t = \begin{cases} 0 & \text{if } |\bar{v}_t| < \epsilon \\ \bar{v}_t & \text{otherwise} \end{cases},  \quad
    \omega_t = \begin{cases} 0 & \text{if } |\bar{\omega}_t| < \epsilon \\ \bar{\omega}_t & \text{otherwise} \end{cases} 
\end{align}
The finalized, mechanically safe velocities $(v_t, \omega_t)$ are then published to the motor drivers and cached as the previous state for the subsequent cycle.

\clearpage
\subsection{State Estimation Comparison}
\label{sup:c3}

\begin{table}[!h]
\centering
\renewcommand{\arraystretch}{1.3}
\setlength{\tabcolsep}{5pt}
\caption{\textbf{State Estimation Resource Overhead.} Additional system resources consumed by the localization nodes evaluated on the NVIDIA Jetson AGX Orin 64\,GB.}
\resizebox{0.8\columnwidth}{!}{%
\begin{tabular}{l c c c}
\hlineB{2.5}
\textbf{Odometry} & \textbf{CPU Usage (\%)} & \textbf{GPU Usage (\%)} & \textbf{Memory (MB)} \\
\hline
ZED VIO & +19.4 & +42.6 & +577.3 \\
Fast-LIVO2 & +26.0 & +13.3 & +367.3 \\
\hlineB{2.5}
\end{tabular}%
}
\vspace{15pt}
\label{sup:tab_10}
\end{table}

\begin{figure*}[!b]
\centering
\includegraphics[width=\textwidth]{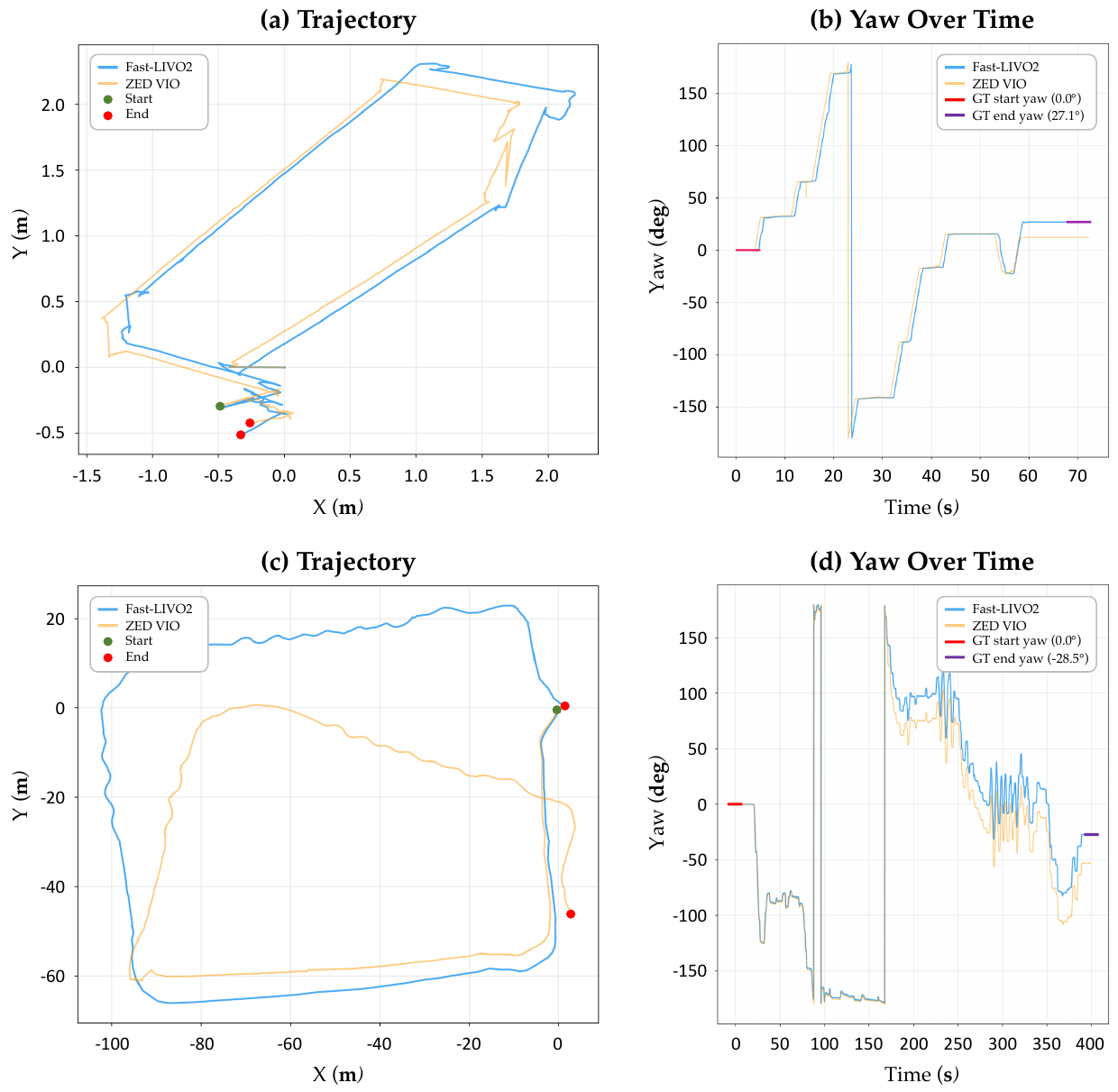}
\caption{
    \textbf{Qualitative Trajectory and Yaw Estimation Analysis.} Representative evaluation runs comparing \textcolor[HTML]{4CAAF5}{\textbf{Fast-LIVO2}} and \textcolor[HTML]{FFCD81}{\textbf{ZED VIO}} in (a, b) indoor and (c, d) outdoor environments. The robot starts at the green dot and ends at the red dot. \textcolor[HTML]{4CAAF5}{\textbf{Fast-LIVO2}} successfully closes the loop in both scenarios, whereas \textcolor[HTML]{FFCD81}{\textbf{ZED VIO}} suffers from severe scale and rotational drift, particularly in the outdoor environment.
}
\label{sup:fig_7}
\end{figure*}

As discussed in Sec. 3.4 and 3.5 of the main paper, our framework requires precise and continuous odometry for two critical operations: spatial trajectory recording for failure recovery and ego-motion compensation for the reactive costmap. While the StereoLabs ZED 2i provides a built-in VIO solution, its reliance on the GPU forces it to share it with the compute-intensive VLA model, resulting in degraded inference speed. 
To resolve this architectural bottleneck, we strategically adopted Fast-LIVO2.By utilizing the onboard 3D LiDAR and the sensor-calibrated ZED 2i, Fast-LIVO2 fuses multimodal data (LiDAR, RGB, IMU) to achieve high precision while executing the primary odometry tasks on the CPU, as detailed in Table~\ref{sup:tab_10}.
The evaluations in Table~\ref{sup:tab_12} and Fig.~\ref{sup:fig_7} highlight the synergistic efficiency of this design. Replacing the default ZED VIO with Fast-LIVO2 protects the VLA model’s computational bandwidth, guaranteeing faster inference without sacrificing LiDAR-enhanced precision. Ultimately, as Table~\ref{sup:tab_11} shows, our fully integrated system (NaVILA, path deviation detection, Fast-LIVO2, and the RL policy) incurs a marginal overhead of only 0.55 s compared to standalone NaVILA. This confirms that strategically distributing odometry to the CPU and utilizing a lightweight RL policy successfully preserves the system's overall real-time performance.


\begin{table}[!b]
\vspace{-10pt}
\centering
\renewcommand{\arraystretch}{1.3}
\setlength{\tabcolsep}{5pt}
\caption{
    \textbf{VLA Inference Time Comparison.} Evaluated over 100 runs each on the NVIDIA Jetson AGX Orin. As verified in Sec.~\ref{sup:b4}, the lightweight RL policy does not affect the VLA's inference speed. While integrating state estimation inevitably adds overhead, Fast-LIVO2 maintains a lower latency than ZED VIO by effectively leveraging the CPU, as demonstrated in Table 2. Consequently, \textbf{Ours} (NaVILA + path deviation detection + Fast-LIVO2 + RL) requires only lightweight scalar operations from the attention heads, introducing a negligible additional overhead of $\sim$0.05\,s.
    Ultimately, with only $\sim$0.55\,s of total additional computation compared to standalone NaVILA, our system successfully executes primary VLA reasoning alongside trajectory recording, path deviation monitoring, and obstacle avoidance.
}
\resizebox{1.0\columnwidth}{!}{%
\begin{tabular}{l c c c c}
\hlineB{2.5}
\textbf{Configuration} & \textbf{Mean (s)} & \textbf{Median (s)} & \textbf{Min / Max (s)} & \textbf{Stdev (s)} \\
\hline
NaVILA & 3.488 & 3.489 & 3.461 / 3.509 & 0.008 \\
\textcolor{gray!50}{NaVILA + RL} & \textcolor{gray!50}{3.488} & \textcolor{gray!50}{3.489} & \textcolor{gray!50}{3.461} / \textcolor{gray!50}{3.509} & \textcolor{gray!50}{0.008} \\
NaVILA + ZED VIO + RL & 4.584 & 4.588 & 4.525 / 4.645 & 0.025 \\
NaVILA + Fast-LIVO2 + RL & 3.971 & 3.968 & 3.918 / 4.046 & 0.024 \\
\hline
\textbf{Ours} & 4.022 & 4.020 & 3.910 / 4.120 & 0.041 \\
\hlineB{2.5}
\end{tabular}%
}
\label{sup:tab_11}
\end{table}

\clearpage

\begin{table*}[!t]
\centering
\renewcommand{\arraystretch}{1.2}
\setlength{\tabcolsep}{5pt}
\caption{
    \textbf{State Estimation Performance Comparison.} Fast-LIVO2 vs. ZED VIO Odometry evaluated over 10 indoor and 10 outdoor runs. The position and yaw errors were measured by computing the accumulated drift relative to a fixed AprilTag before and after each run. The best (lowest) error in each metric is highlighted in bold.
}
\resizebox{\textwidth}{!}{%
\begin{tabular}{l l c c c c c c}
\hlineB{2.5}
\multirow{2}{*}{\textbf{Environment}} & \multirow{2}{*}{\textbf{Method}} & \multicolumn{2}{c}{\textbf{GT Pose Error (m)}} & \multicolumn{2}{c}{\textbf{GT Yaw Error ($^\circ$)}} & \multicolumn{2}{c}{\textbf{Static Drift (cm)}} \\
\cmidrule(lr){3-4} \cmidrule(lr){5-6} \cmidrule(lr){7-8}
& & \textbf{Median} & \textbf{Mean} & \textbf{Median} & \textbf{Mean} & \textbf{Median} & \textbf{Mean} \\
\hline
\multirow{2}{*}{\textbf{Indoor}} 
& Fast-LIVO2 & \textbf{0.082} & \textbf{0.331} & \textbf{0.640} & \textbf{4.463} & 0.080 & 0.080 \\
& ZED VIO & 0.276 & 0.413 & 1.345 & 5.850 & \textbf{0.025} & \textbf{0.053} \\
\hline
\multirow{2}{*}{\textbf{Outdoor}} 
& Fast-LIVO2 & \textbf{0.375} & \textbf{0.731} & \textbf{6.690} & \textbf{13.469} & \textbf{0.065} & 0.166 \\
& ZED VIO & 0.839 & 4.589 & 8.560 & 22.314 & 0.070 & \textbf{0.124} \\
\hlineB{2.5}
\end{tabular}%
}
\vspace{5pt}
\label{sup:tab_12}
\end{table*}

\clearpage



\section{Real-World Deployment Results}
\label{sup:d}

\begin{figure*}[!b]
\vspace{-20pt}
\centering
\includegraphics[width=0.98\textwidth]{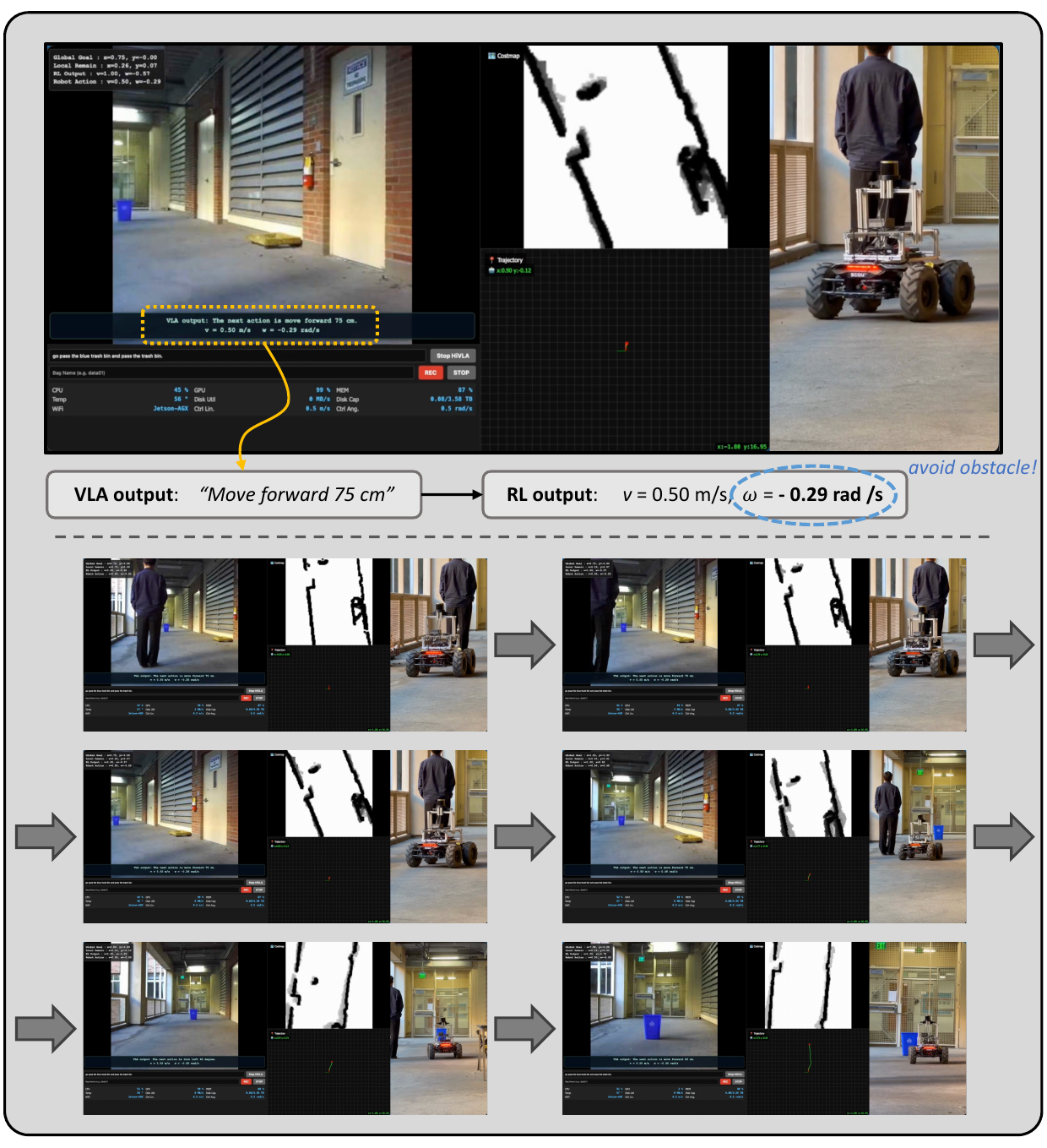}
\vspace{-5pt}
\caption{
    \textbf{Robust Obstacle Avoidance via Low-Level RL Policy.} This sequence demonstrates the system's ability to correct unsafe high-level commands in real-time. Although the VLA model naively outputs a forward movement command ("Move forward 75 cm") despite a pedestrian blocking the path, our RL policy independently detects the obstacle. It immediately overrides the naive command by executing a safe avoidance maneuver (v=0.50 m/s, ω=−0.29 rad/s), allowing the robot to steer around the pedestrian without collision and continue its navigation.
}
\vspace{-10pt}
\label{sup:fig_8}
\end{figure*}

\begin{figure*}[!t]
\centering
\includegraphics[width=\textwidth]{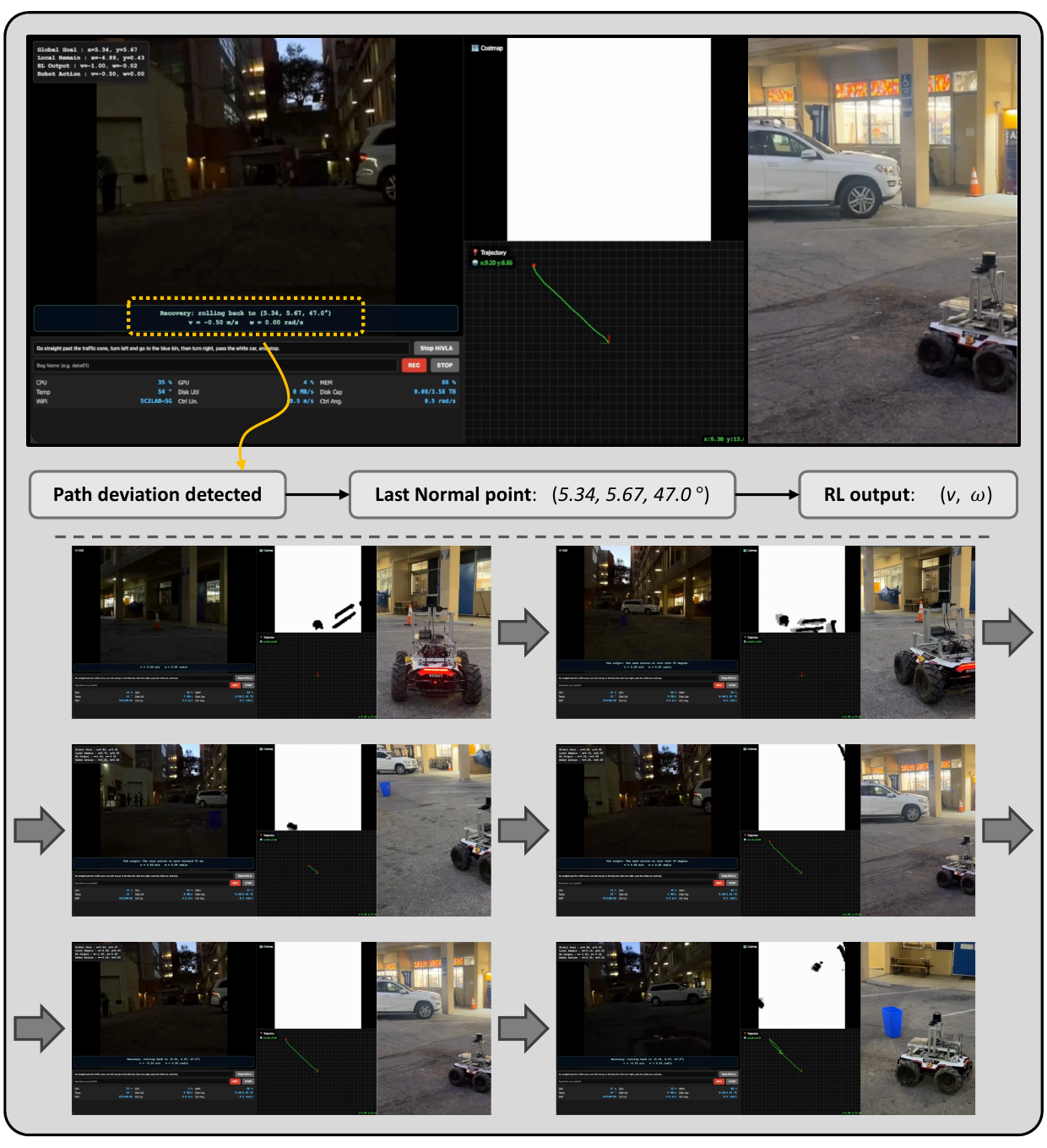}
\caption{
    \textbf{Path Deviation Detection and Autonomous Trajectory Recovery.} 
    This sequence demonstrates the system's rollback mechanism. When a path deviation caused by VLA hallucination is detected during navigation, the system aborts the erroneous trajectory. It then initiates a recovery phase toward the last normal state (e.g., x=5.34, y=5.67, z=47.0°) from the real-time recorded trajectory. During this process, the robot automatically returns to that point via the shortest path, guided by the obstacle avoidance actions of the RL policy, ensuring the mission resumes smoothly.
}
\label{sup:fig_9}
\end{figure*}

This section presents real-world testing of our fully integrated framework, combining the VLA backbone with path deviation detection, a rollback mechanism to the saved normal trajectory, and a low-level RL avoidance policy. We highlight two key capabilities. First, Fig.~\ref{sup:fig_8} demonstrates robust, real-time obstacle avoidance despite naive VLA commands. Second, Fig.~\ref{sup:fig_9} illustrates autonomous trajectory recovery to the last normal state following a path deviation.

